\documentclass{article}

\PassOptionsToPackage{numbers, compress}{natbib}

\usepackage[preprint]{neurips_2023}

\usepackage{amsmath,amsfonts,bm}

\def\eqref#1{equation~\ref{#1}}

\def\1{\bm{1}}

\def\rvb{{\mathbf{b}}}

\def\rvh{{\mathbf{h}}}

\def\rvp{{\mathbf{p}}}

\def\rvw{{\mathbf{w}}}

\def\rvz{{\mathbf{z}}}

\def\rmH{{\mathbf{H}}}

\def\rmW{{\mathbf{W}}}

\def\rmZ{{\mathbf{Z}}}

\def\vx{{\bm{x}}}
\def\vy{{\bm{y}}}

\def\evb{{b}}

\def\evp{{p}}

\def\evy{{y}}
\def\evz{{z}}

\def\mM{{\bm{M}}}

\def\mX{{\bm{X}}}
\def\mY{{\bm{Y}}}

\DeclareMathAlphabet{\mathsfit}{\encodingdefault}{\sfdefault}{m}{sl}
\SetMathAlphabet{\mathsfit}{bold}{\encodingdefault}{\sfdefault}{bx}{n}

\def\sD{{\mathbb{D}}}

\def\sK{{\mathbb{K}}}

\def\sR{{\mathbb{R}}}

\def\emM{{M}}

\def\emZ{{Z}}

\usepackage{hyperref}
\hypersetup{
    colorlinks=true,
    linkcolor=[rgb]{0.84,0.15,0.16},
    citecolor=[rgb]{0.12,0.47,0.71},
}

\usepackage{url}

\usepackage{times}
\usepackage{helvet}
\usepackage{courier}
\usepackage{nicefrac}       
\usepackage{microtype}      
\usepackage{xcolor}         
\usepackage{color,colortbl}
\usepackage{caption}
\definecolor{myred}{rgb}{0.84,0.15,0.16}
\definecolor{myblue}{rgb}{0.12,0.47,0.71}

\usepackage{enumitem}

\usepackage{graphicx}
\usepackage{caption}
\usepackage{subcaption}
\usepackage{wrapfig}

\usepackage{booktabs}       
\usepackage{tabularx}
\usepackage{array}
\usepackage{multirow}
\usepackage{siunitx}
\usepackage{arydshln}
\usepackage{amssymb} 
\usepackage{dsfont} 

\usepackage{amsthm} 

\usepackage{algorithm}
\usepackage{algpseudocode}

\usepackage{soul}

\title{Revisiting Softmax Masking: \\Stop Gradient for Enhancing Stability in Replay-based Continual Learning}

\author{%
  Hoyong Kim\textsuperscript{1}, 
  Minchan Kwon\textsuperscript{2}, 
  Kangil Kim\textsuperscript{1}\thanks{corresponding author}\\
  Artificial Intelligence Graduate School\textsuperscript{1},
  Graduate School of AI\textsuperscript{2}
  \\
  Gwangju Institute of Science and Technology\textsuperscript{1}, 
  KAIST\textsuperscript{2} \\
  Republic of Korea \\
  \texttt{hoyong.kim.21@gm.gist.ac.kr, \{kmc020700, kangil.kim.01\}@gmail.com} \\
}

\begin{document}

\maketitle

\begin{abstract}
In replay-based methods for continual learning, replaying input samples in episodic memory has shown its effectiveness in alleviating catastrophic forgetting.
However, the potential key factor of cross-entropy loss with softmax in causing catastrophic forgetting has been underexplored.
In this paper, we analyze the effect of softmax and revisit softmax masking with negative infinity to shed light on its ability to mitigate catastrophic forgetting. 
Based on the analyses, it is found that negative infinity masked softmax is not always compatible with dark knowledge. 
To improve the compatibility, we propose a \textit{general masked softmax} that controls the stability by adjusting the gradient scale to old and new classes. 
We demonstrate that utilizing our method on other replay-based methods results in better performance, primarily by enhancing model stability in continual learning benchmarks, even when the buffer size is set to an extremely small value.
\end{abstract}

%
%
\section{Introduction}
In continual learning, catastrophic forgetting is a challenging problem. 
It refers to the difficulty of preserving previously trained information in a model when learning from future task data~\cite{catastrophic_forgetting_origin}.  
Recent works have focused on utilizing episodic memory to stably transfer information, allowing for accurate recall of previous task information without sacrificing adaptation on current tasks~\cite{cil_replay_a-gem,cil_replay_bic,cil_replay_derpp,cil_replay_dual,cil_replay_er,cil_replay_ER-ACE,cil_replay_gdumb,cil_replay_gem,cil_replay_icarl,cil_replay_lucir,cil_replay_rpc,cil_replay_select,cil_replay_stream,cil_replay_xder}.  
This feature is also known as stability~\cite{catastrophic_forgetting_remem}.
For instance, a well-performing model~\cite{cil_replay_derpp} stores real samples along with their confidence outputs at each update iteration, while balancing their size across tasks. 
However, this methodology does not directly modify the training dynamics of a neural network. 
In particular, the gradient from cross-entropy loss with softmax is one of the main factors that can cause catastrophic forgetting, but it has not yet been extensively studied.

\begin{figure}[t]
\centering
    \centering
    \includegraphics[width=0.95\linewidth]{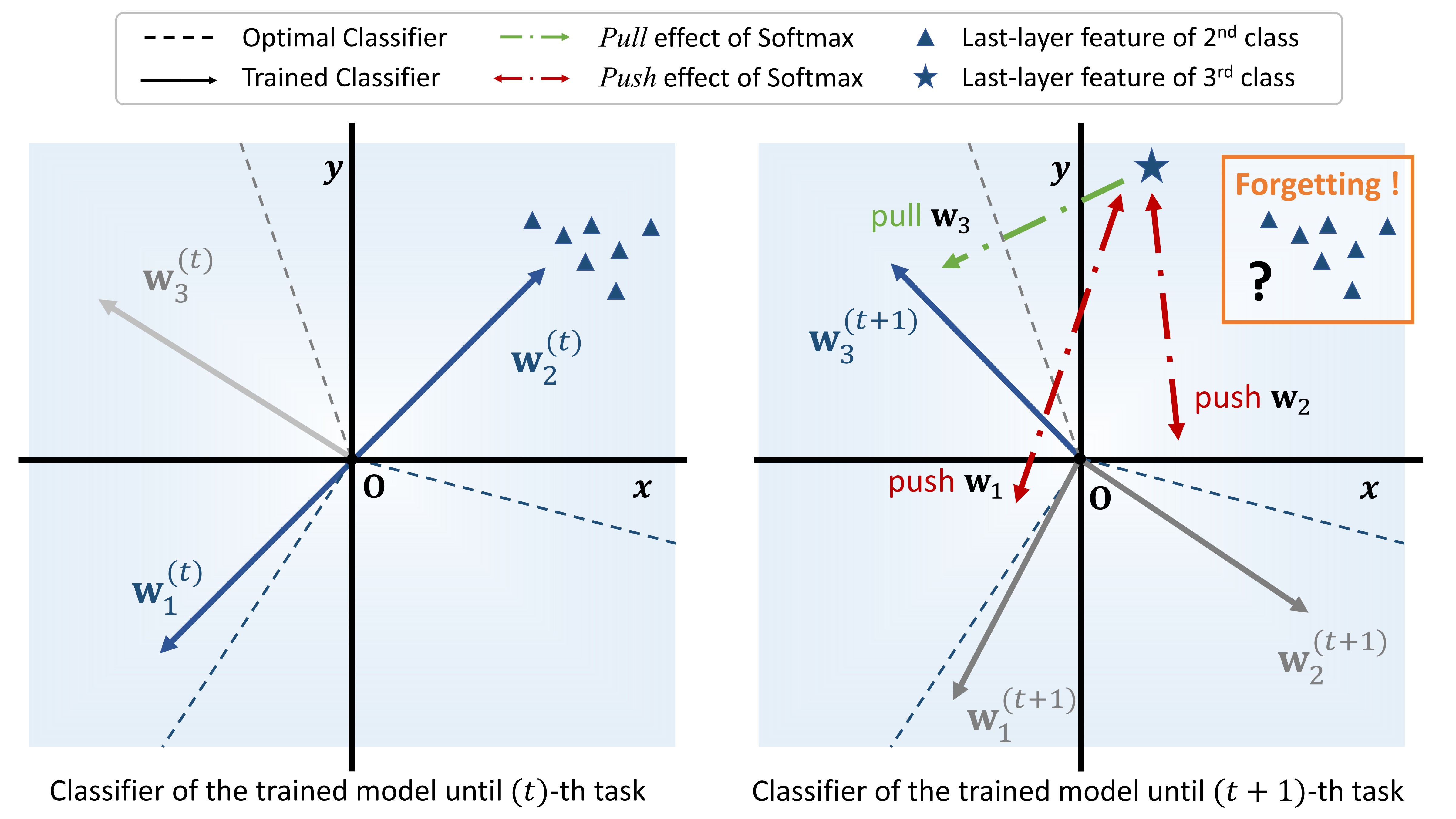}
\caption{Relation between Catastrophic Forgetting and Cross-Entropy Loss with Softmax in Continual Learning.}
\label{fig:main_problem}
\end{figure}

In numerous works~\cite{softmax_overconfidence_logitnorm,softmax_overconfidence_text_clf,softmax_overconfidence_medical,softmax_overconfidence_recommend,softmax_ddu}, the use of cross-entropy loss with softmax results in the model being overconfident in its pretrained knowledge. 
This is achieved by increasing the confidence of a target class while decreasing the confidence of other classes. 
The \textit{pull} and \textit{push} effects~\cite{softmax_pull_push} exacerbate catastrophic forgetting in continual learning. 
This occurs because the gradients towards target classes in the current task pull last-layer features towards their respective class weight vectors while pushing other class weight vectors in previous tasks, without any access to input samples of those tasks, as illustrated in~\autoref{fig:main_problem}.
In terms of the \textit{push} effect of softmax, prior replay-based methods have alleviated it by replaying the input samples of previous tasks in external memory and retraining their respective class weight vectors. 
However, the prior replay-based method of using the re-aligning features and class weight vectors implicitly relieves the \textit{push} effect of softmax.

To explicitly control the \textit{pull} and \textit{push} effect of softmax, we revisit the common masked softmax and propose using a \textit{general masked softmax} approach. 
While masking is a widely used technique in literature, its implications for model stability have been under-investigated. 
By using masked softmax, we can ensure stability while still achieving our desired results. 
This approach utilizes negative infinity value or any real value to mask probability for the classes in previous and future tasks, effectively enhancing stability. 
In this study, we empirically analyze the confidence change, plasticity (overall accuracy across tasks), and stability (accuracy of previous tasks at each training step) of split datasets in class and task incremental continual learning. 
We vary the memory size in replay-based methods and demonstrate that masking in softmax is effective in adjusting model stability. 
Our results show that replay-based methods with general masked softmax outperform various datasets in continual learning scenarios.

Our contribution is:
\begin{itemize}
    \item  We revisit the use of softmax masking and highlight its ability to preserve previously trained confidence information in continual learning. 
    \item We propose a \textit{general masked softmax} approach and demonstrate its effectiveness in utilizing the confidence holding, even with an extremely low buffer size. 
    \item Empirical results show that our approach enhances stability on well-known replay-based continual learning models and benchmarks while maintaining sufficiently large plasticity. 
\end{itemize}

%
%
\section{Related Work}
\label{section: related work}

\paragraph{Cross-entropy with Softmax in Replay-based Continual Learning.}
Similar to common image classification settings, replay-based continual learning employs cross-entropy loss with softmax. 
Experience Replay (ER)~\cite{cil_replay_er} uses buffer samples in episodic memory and it has the effect of offsetting the catastrophic forgetting to some extent by canceling out the \textit{push} effect of current samples via the gradients from these samples. 
GEM~\cite{cil_replay_gem} and A-GEM~\cite{cil_replay_a-gem} utilize inequality constraints to prevent an increase in the gradients of previous tasks while allowing their decrease through the storage and replay of these gradients in episodic memory. 
In addition to buffer samples, some prior works have employed a distillation loss term to retain previous knowledge. 
iCaRL~\cite{cil_replay_icarl} trains images from a current task and exemplars from previous tasks using cross-entropy and distillation loss. 
The exemplars are updated and stored in episodic memory. 
Dark Experience Replay (DER) and DER++~\cite{cil_replay_derpp} store the logits of previous tasks in episodic memory and use them for distillation loss in the current task. 
These replay-based methods are simple but effective in preventing catastrophic forgetting by recovering buffer samples and, in some cases, their logits from previous tasks. 
However, while the losses related to buffer samples cancel out the \textit{push} effect of softmax on current samples, they do not consider the \textit{push} and \textit{pull} effects of softmax and do not directly control them.

Some replay-based methods have focused on the classifier to alleviate the forgetting caused by the \textit{push} effect~\cite{cil_replay_rpc,softmax_pull_push}. 
One such method is Regular Polytope Classifier (RPC)~\cite{cil_replay_rpc}, which fixes the classifier of models as regular polytope shapes to escape the influence of the \textit{push} effect. 
Another method~\cite{softmax_pull_push}, inspired by neural collapse, also fixes the classifier but in a different shape - a simplex equiangular tight frame - which occurs in post-training after zero training error. 
However, neither of these methods directly controls the softmax. 
In contrast, the use of softmax masking is a simple and direct method to prevent the \textit{push} effect of softmax. It is effective in alleviating catastrophic forgetting by increasing the stability of models in continual learning scenarios.

\paragraph{Masked Softmax in Continual Learning.}
Although the benefits of softmax masking in improving continual learning are recognized, the exact mechanisms behind these benefits are not well understood. 
There are various approaches to continual learning~\cite{van2022three}, but our paper focuses on the scenario of task and class incremental continual learning in offline, where the model learns new classes progressively with or without task ids~\cite{cil_replay_derpp,cil_replay_xder,cha2021co2l}. 
In the scratch-initiated setting, researchers have explored the use of negative infinity masking on the softmax function to enhance performance~\cite{caccia2022new}. 
Additionally, their findings were limited to an online environment where data is accessed only once. 
Many recent studies use visual prompts to mitigate forgetting in pre-trained models by employing softmax masking~\cite{wang2022learning,Smith_2023_CVPR,wang2022dualprompt}.
Previous studies have shown that visual prompts often rely on the masking approach. 
The softmax masking technique, even when used alone, can produce impressive results with large pre-trained models~\cite{Smith_2023_CVPR}. 
Building on prior research on softmax masking, we revisit the softmax masking approach and expand it into a general form. 
We then integrate the general masked softmax with a state-of-the-art continual learning framework, starting from scratch, and explore its implications.

%
%
\begin{figure}[t]
\centering
\begin{subfigure}[b]{0.32\textwidth}
    \centering
    \includegraphics[width=\linewidth]{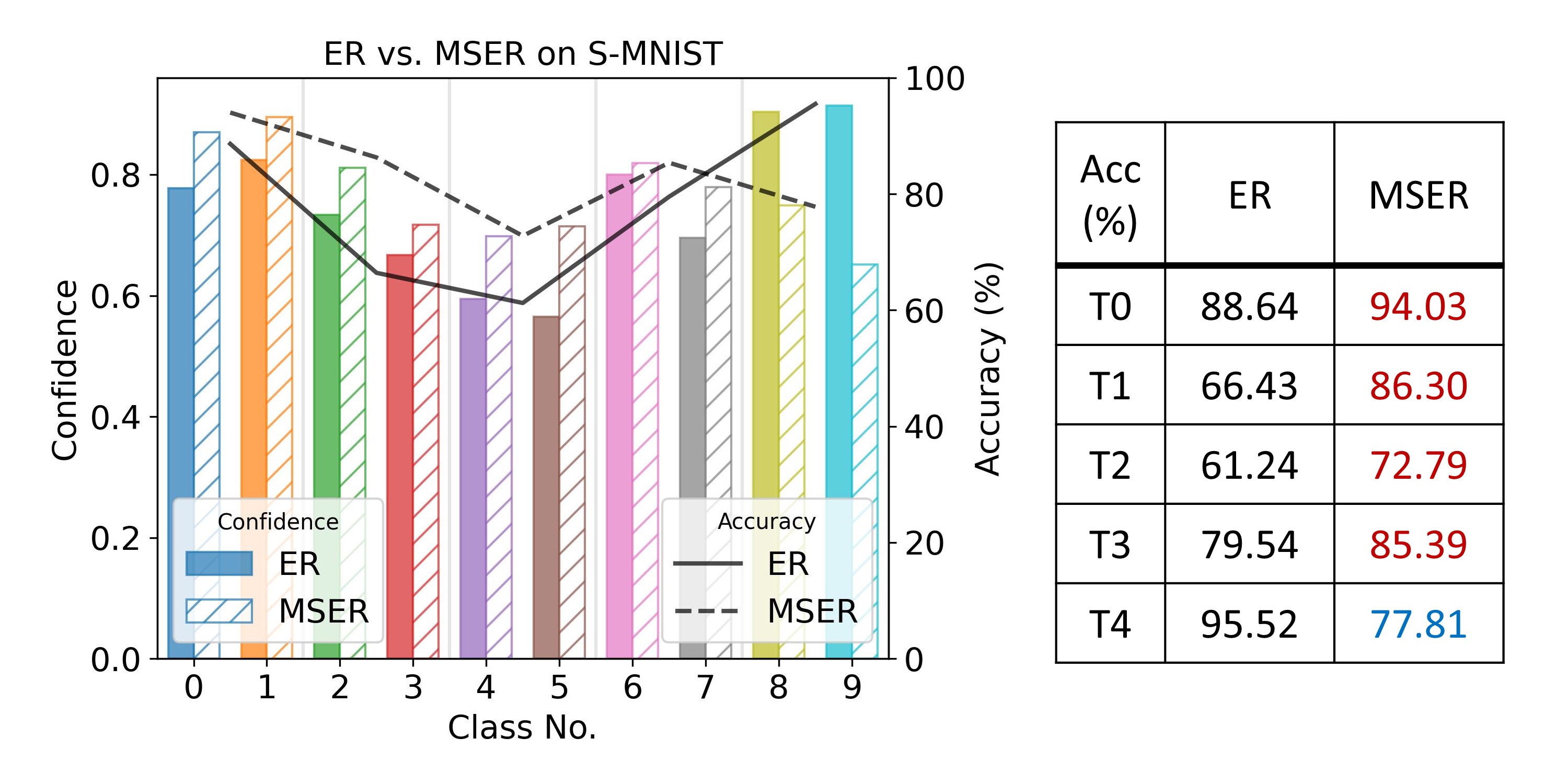}
    \captionsetup{justification=centering}
    \caption{{\scriptsize Confidence and Accuracy on S-MNIST \\ (Experiments w/o distillation)}}
    \label{fig:analysis:er-mnist-allcls}
\end{subfigure}
\begin{subfigure}[b]{0.32\linewidth}
    \centering
    \includegraphics[width=\linewidth]{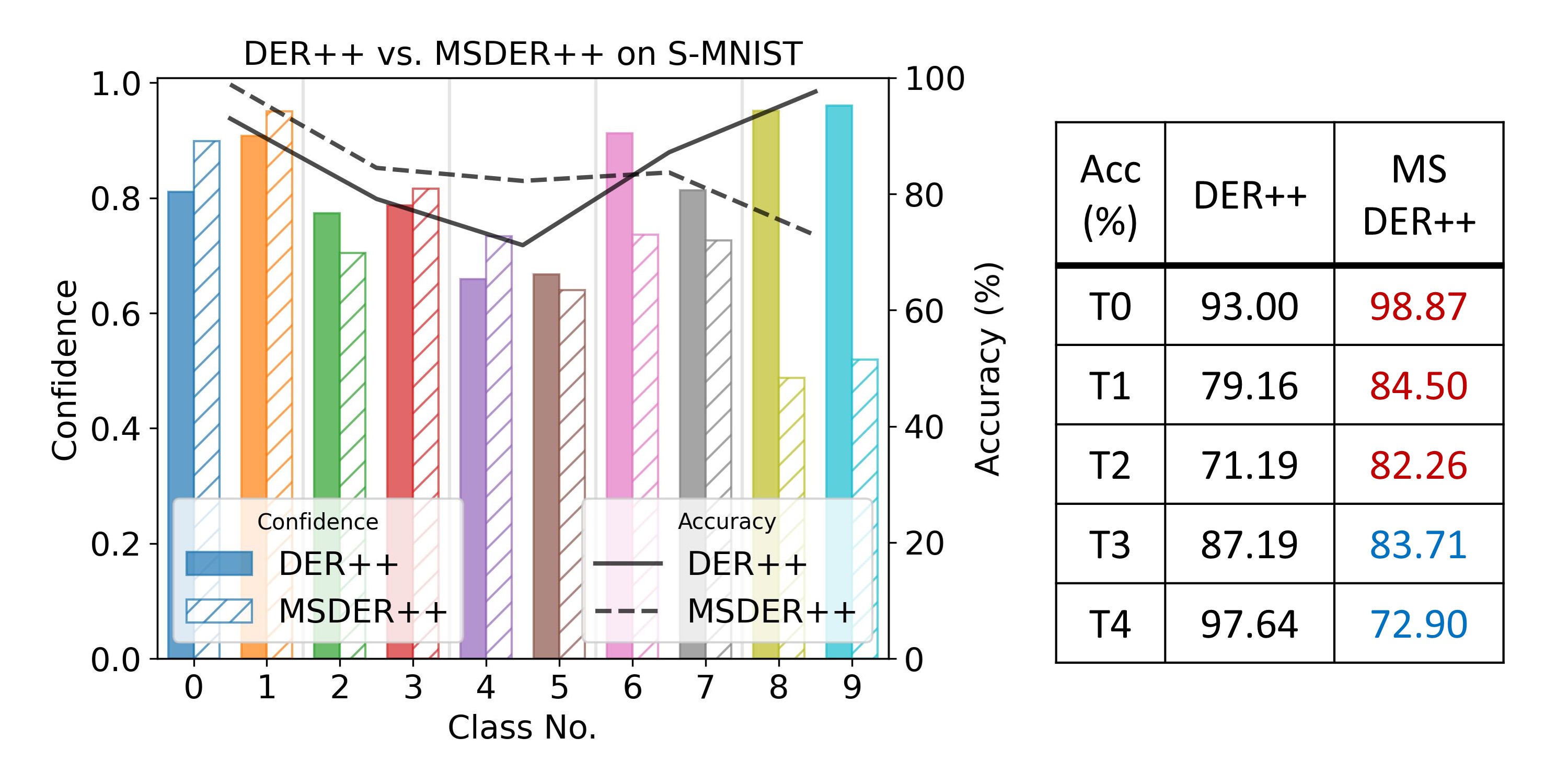}
    \captionsetup{justification=centering}
    \caption{{\scriptsize Confidence and Accuracy on S-MNIST \\ (Experiments w/ distillation)}}
    \label{fig:analysis:derpp-mnist-allcls}
\end{subfigure}
\begin{subfigure}[b]{0.32\linewidth}
    \centering
    \includegraphics[width=\linewidth]{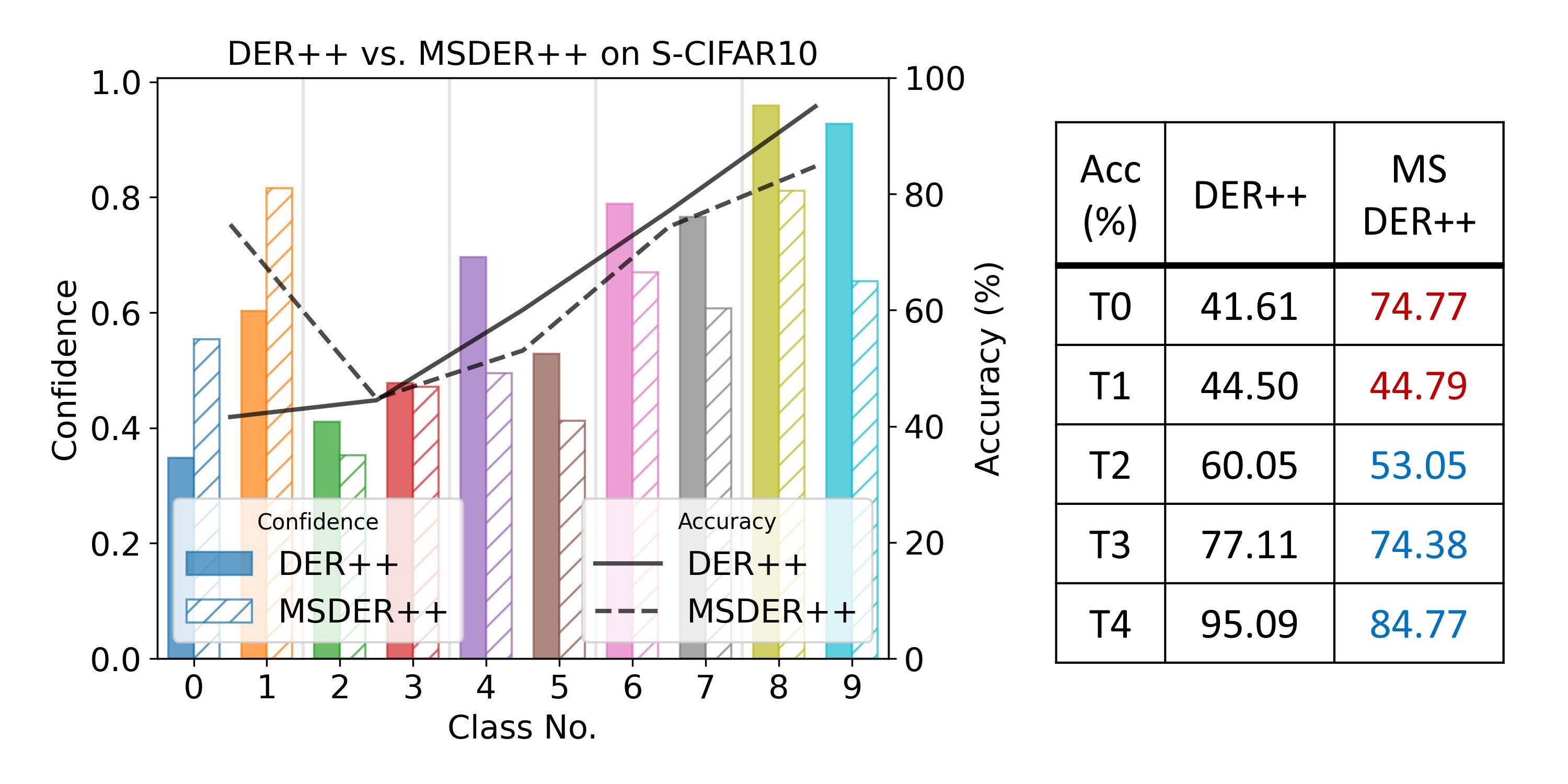}
    \captionsetup{justification=centering}
    \caption{{\scriptsize Confidence and Accuracy on S-CIFAR10 \\ (Experiments w/ distillation)}}
    \label{fig:analysis:derpp-cifar10-allcls}
\end{subfigure}
\begin{subfigure}[b]{0.32\linewidth}
    \centering
    \includegraphics[width=\linewidth]{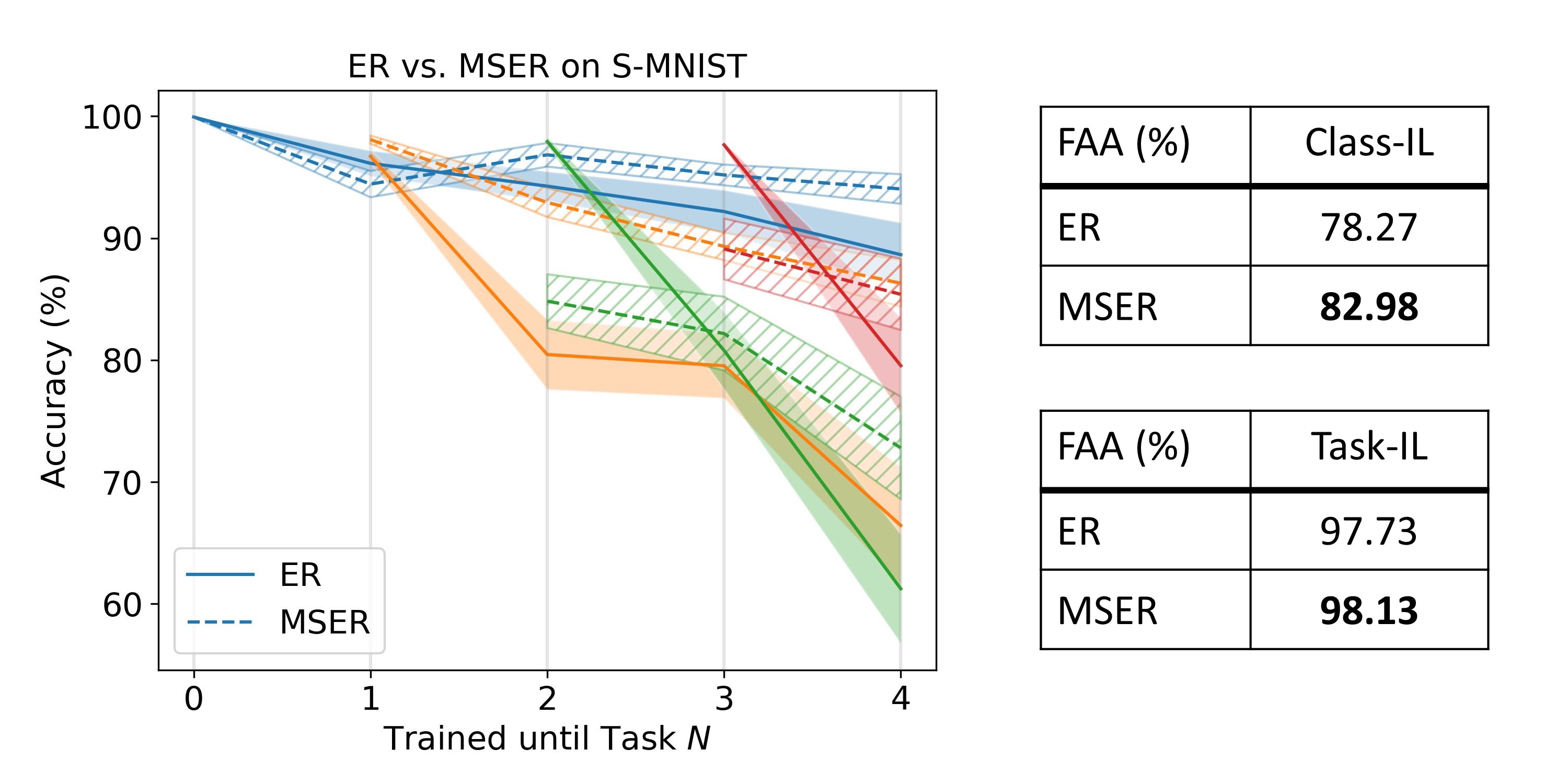}
    \captionsetup{justification=centering}
    \caption{{\scriptsize Change in Accuracy on S-MNIST \\ (Experiments w/o distillation)}}
    \label{fig:analysis:er-mnist-changeacc}
\end{subfigure}
\begin{subfigure}[b]{0.32\linewidth}
    \centering
    \includegraphics[width=\linewidth]{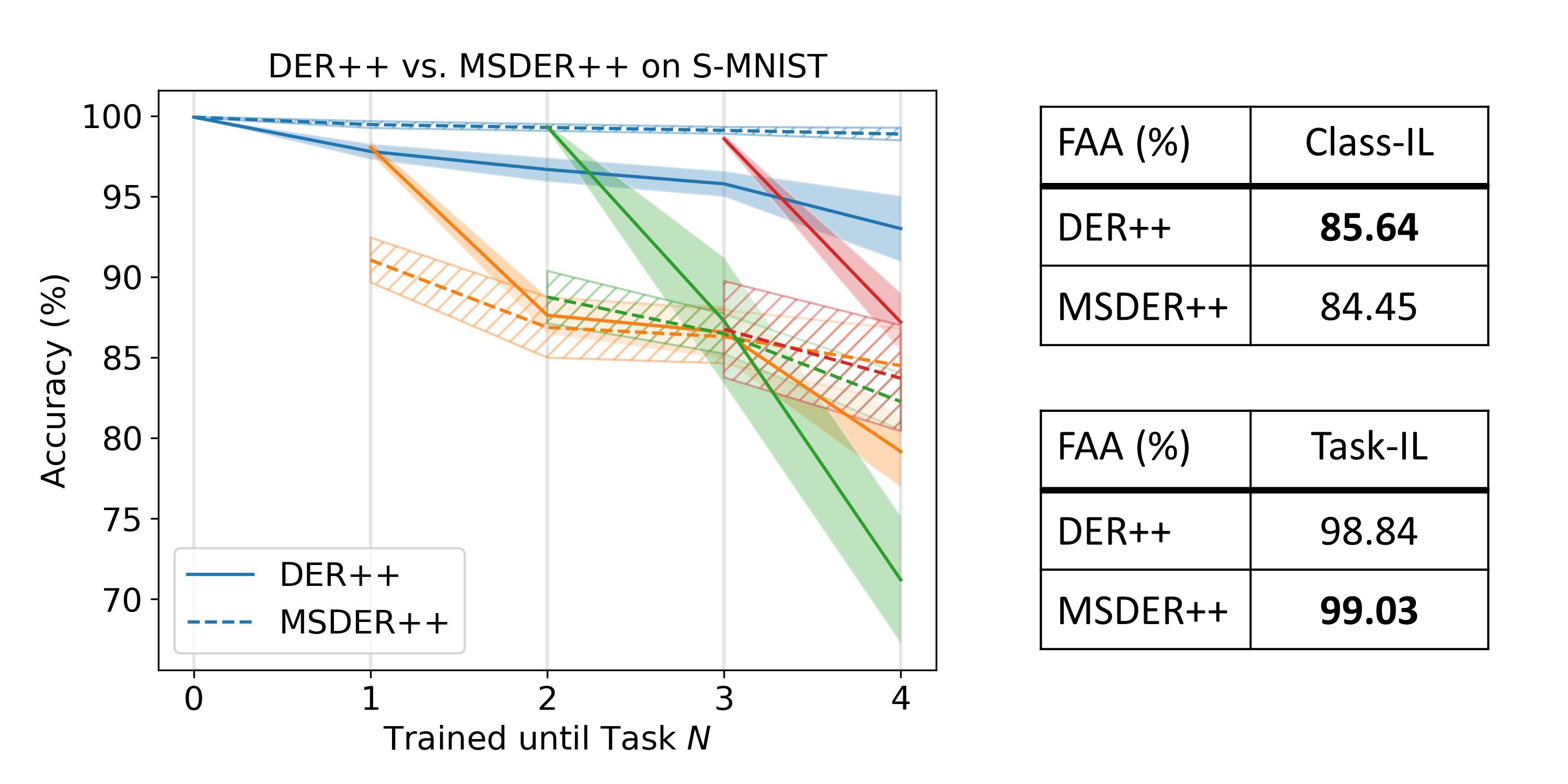}
    \captionsetup{justification=centering}
    \caption{{\scriptsize Change in Accuracy on S-MNIST \\ (Experiments w/ distillation)}}
    \label{fig:analysis:derpp-mnist-changeacc}
\end{subfigure}
\begin{subfigure}[b]{0.32\linewidth}
    \centering
    \includegraphics[width=\linewidth]{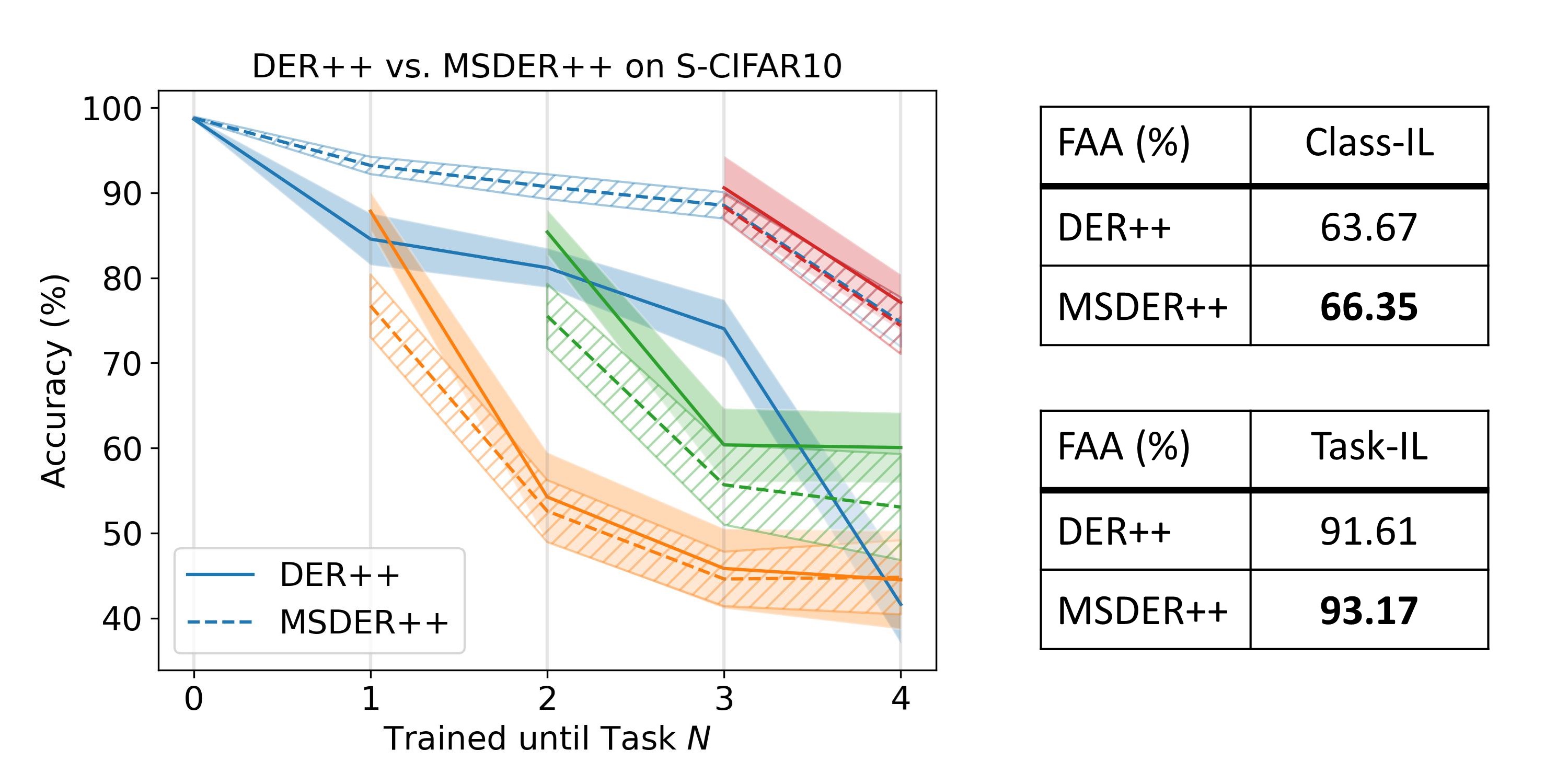}
    \captionsetup{justification=centering}
    \caption{{\scriptsize Change in Accuracy on S-CIFAR10 \\ (Experiments w/ distillation)}}
    \label{fig:analysis:derpp-cifar10-changeacc}
\end{subfigure}
\caption{
Comparison Experiments between baseline and masked softmax in class-incremental learning. 
(1\textsuperscript{st} row) Confidence of each classes and accuracy of each tasks after the final task. 
Confidence is measured by the output of the softmax function from the logits of input samples.
(2\textsuperscript{nd} row) Change in average accuracy for each task as the models are trained up to the $N$-th task.
Task-wise average accuracy is defined as the mean of accuracies of each class in the task.
The dataset and methods used in each experiment are described in the title and caption.
(Acc: task-wise average test accuracy, T$n$: $n$-th task, {\color{myred} red}: higher than baseline, {\color{myblue} blue}: lower than baseline, FAA: Final Average Accuracy, Class-IL: Class-Incremental Learning, Task-IL: Task-Incremental Learning, \textbf{bold}: best FAA)
}
\label{fig:analysis}
\end{figure}

\section{Effects of Softmax Masking in Continual Learning}
\label{sec:effects_of_ms}

\paragraph{Continual Learning Setting.}
Let $\sD = {(\mX, \mY)}$ datasets with $N$ pairs of input samples $\mX$ and one-hot encoding vectors $\mY$ associated with $\mX$ where the element at the input sample's class label is 1 and the others are 0.
When there is a class label set $\sK = \{1, 2, \dots, K\}$ and $T$ tasks, we denote $N_k^{(t)}$ as the number of pairs $(\mX_k^{(t)}, \mY_k^{(t)})$ in $k$-th class belong to $t$-th task, \textit{i.e.}, $N^{(t)}=\sum_{k \in \sK^{(t)}} N_k^{(t)}$, where $\sK^{(t)}$ is a set of class labels in the $t$-th task.

\subsection{Analyses on Masked Softmax in Continual Learning}

\paragraph{Change in Confidence and Task-wise Average Accuracy of Model.}
In common masking methods, they set the masking value to negative infinity and multiply it to logits from input samples before softmax.
Thus, we also use a negative infinity mask to implement masked softmax in continual learning setting, \textit{i.e.},
\begin{align}
\begin{split}
    \mM^{(t)} &= (\emM_{i,j})_{1 \leq i \leq N^{(t)}, 1 \leq j \leq K}, \\
    \rvp^{(t)} &= \textsc{Softmax}(\mM^{(t)} \odot (\rmW^{\intercal} \rmH^{(t)} + \rvb \textbf{1}_{N^{(t)}}^{\intercal})),
\end{split}    
\end{align}
where $\mM$ is a negative infinity mask, $\emM_{i,j} = 1~\text{if}~j \in \sK^{(t)}~\text{otherwise}~-\infty$, $\rmH$ is last-layer features of input samples, and $\rvp$ is their confidence interpreted to probabilistic distribution by softmax function.
$\rmW \in \sR^{D \times K}$ and $\rvb \in \sR^{K}$ are weight parameters of the classifier, $\textbf{1}_n \in \sR^{n}$ is the ones vector, and $\odot$ is the Hadamard product.

Upon the definition of masked softmax, we conducted comparison experiments between ER and ER with masked softmax (MSER) and visualized the results, as shown in~\autoref{fig:analysis:er-mnist-allcls} and~\autoref{fig:analysis:er-mnist-changeacc}.
In~\autoref{fig:analysis:er-mnist-allcls}, the use of masked softmax results in increased confidences of classes in the initial tasks, while the confidences of others in recent tasks decreased slightly. 
This indicates that setting the masked softmax to a negative infinity value is an effective way to maintain confidence in old classes by preventing gradients from flowing towards them. 
The~\autoref{fig:analysis:er-mnist-changeacc} also supports the effect of masked softmax with a negative infinity value. 
It shows that the initial accuracy of each class in the task is inferior, but it maintains better than before. 
In addition, MSER has demonstrated their superior final average accuracy in both class-incremental learning (ER: 78.27\% $<$ MSER: 82.98\%) and task-incremental learning (ER: 97.73\% $<$ MSER: 98.33\%). 
Therefore, we conclude that masking a negative infinity value to the logits of old and new classes can improve the model performance by increasing stability of models.

\paragraph{Masked Softmax with Dark Knowledge.}
However, an increment in stability does not always work result in improved model. 
For instance, when the logits of input samples from old classes were trained with negative infinity masked softmax and then transferred to the future tasks, the model exhibited inferior performance compared to its previous performance, as illustrated in~\autoref{fig:analysis:derpp-mnist-allcls} and~\autoref{fig:analysis:derpp-mnist-changeacc}. 
In~\autoref{fig:analysis:derpp-mnist-allcls}, the confidence of classes in the initial tasks are well preserved after training all tasks, while the confidences of posterior tasks are lower than DER++. 
This phenomenon is likely caused by transferring wrongly trained logits with negative infinity masked softmax and distilling them to the model training on the current task. 
In addition, the model trains new input samples of the current task with negative infinity masked softmax, which ensures the class weight vectors of old classes remain unchanged. 
As a result, the realignment of the class weight vectors of old classes with current knowledge is solely dependent on buffer samples. 
However, these samples are restricted by the size of episodic memory, which can lead to overfitting and finally result in a low inductive bias about previous tasks. 
It means that extremely increasing the model stability from negative infinity masked softmax is not always beneficial for continual learning to achieve high performance. 
This is because it can hinder the model's ability to adapt new knowledge from future tasks. 
The~\autoref{fig:analysis:derpp-cifar10-changeacc} demonstrates this trend, where the initial average accuracy of each task is lower than DER++. 
To the best of our knowledge, the model only has superior performance with masked softmax in class-incremental and task-incremental learning if the improvement of stability increment is larger than the decrement of plasticity, as compared to the results in~\autoref{fig:analysis:derpp-mnist-changeacc} and~\autoref{fig:analysis:derpp-cifar10-changeacc}.

\paragraph{Motivation.}
The research question we address is \textit{how to control the trade-off between stability and plasticity in masked softmax}.
To answer this question, we propose a general masked softmax that replaces the logits of old and new classes in previous and future tasks with not only a negative infinity but also any real values, thus controlling the \textit{push} effect of softmax.
We introduce our approach with explanation in terms of gradients (\autoref{subsec:gms}), demonstrate the effectiveness of our method in comparison experiments with ER and DER++ on continual learning benchmarks (\autoref{subsec:cl_benchmarks}), even in extremely low buffer size environment (\autoref{subsec:cl_lowbuf}), and lastly studies about how the stability is changed as the masking values (\autoref{subsec:ablation}).

\begin{wrapfigure}{l}{0.5\linewidth}
\centering
    \includegraphics[width=\linewidth]{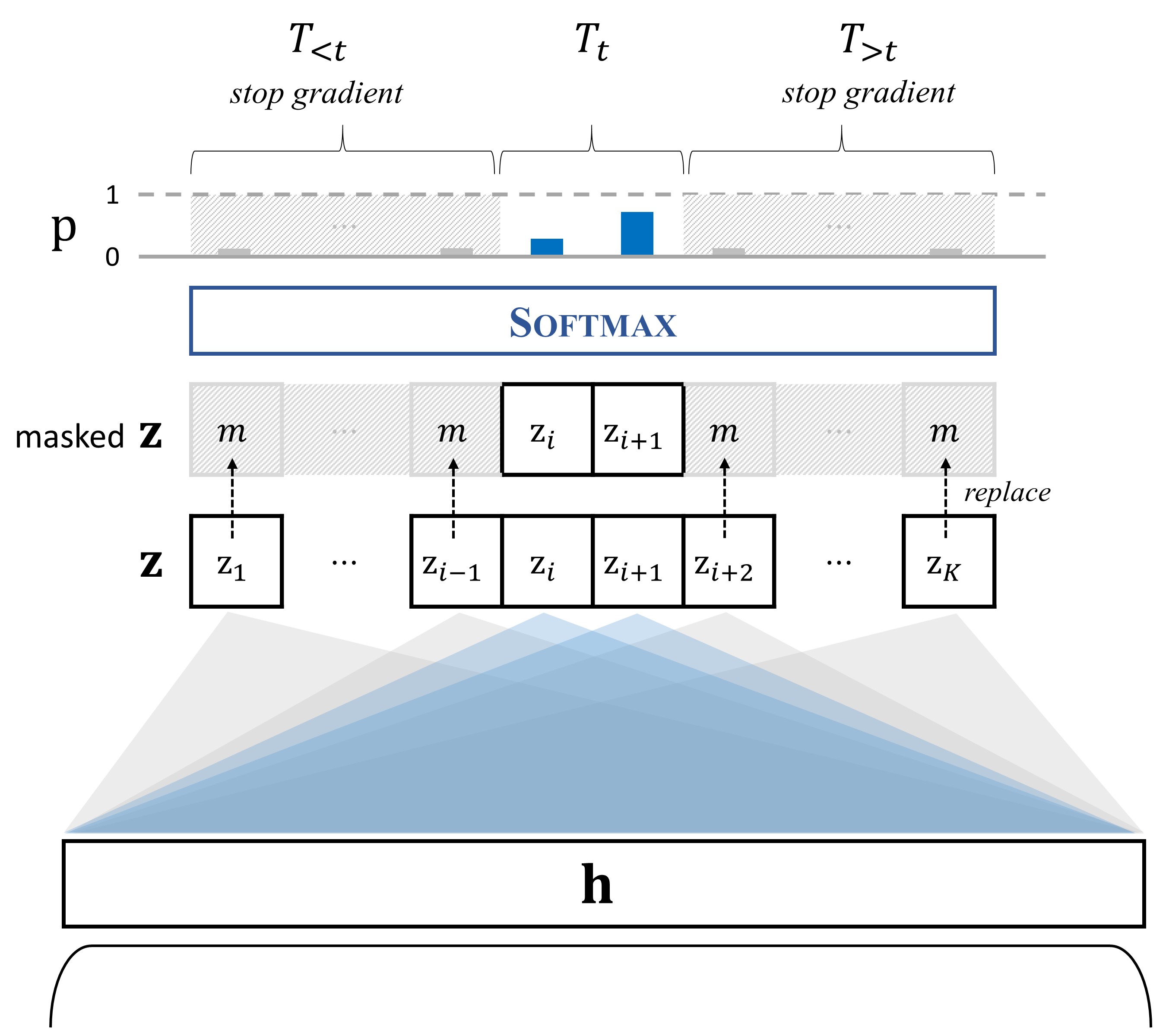}
\caption{
Overview of general masked softmax. 
First, make masked logits $\rvz$ by replacing the logits of old and new classes to a masking value $m$. 
Second, apply softmax function on the masked logits and achieve their confidence $\rvp$. 
Lastly, backward the loss while stop-gradient on replaced logits.}
\label{fig:overview}
\vspace{-70pt}
\end{wrapfigure}

\subsection{General Masked Softmax}
\label{subsec:gms}

In the same setting to~\autoref{sec:effects_of_ms}, we can define \textit{general masked softmax} at $t$-th task as illustrated in~\autoref{fig:overview}:
\begin{align}
\begin{split}
    \rmZ^{(t)} &= \rmW^{\intercal} \rmH^{(t)} + \rvb \textbf{1}_{N^{(t)}}^{\intercal} \\
    \rmZ^{(t)'} &= (\emZ'_{i,j})_{1 \leq i \leq N^{(t)}, 1 \leq j \leq K}, \\
    \rvp^{(t)} &= \textsc{Softmax}(\rmZ^{(t)'}),
\end{split}    
\end{align}
where $\rmZ^{(t)'}$ is a masked logits, $\emZ'_{i,j} = \emZ_{i,j}~\text{if}~j \in \sK^{(t)}~\text{otherwise}~m \in [\text{-}\infty,0]$.

\paragraph{Inter-task Interference.}
The softmax function is a probabilistic activation function that ensures all elements in the output vector are positive and their sum equals 1, \textit{i.e.}, 
\begin{equation}
    \evp_i = \frac{\exp \left( \evz_i \right)}{\sum_{j=1}^{K} \exp \left( \evz_j \right)},
\end{equation}
where $z_i = \rvw_i^{\intercal} \rvh + \evb_i$.

In classification models utilizing cross-entropy loss with softmax, the cost function and gradient are defined as:
\begin{align}
\begin{split}
    \mathcal{L}_{\text{SCE}} \left( \vx, \vy \right) 
    &= -\sum_{i=1}^{K} \evy_i \log \left( \evp_i \right) \\
    \frac{\partial \mathcal{L}_{\text{SCE}}}{\partial \rvz}
    &= \rvp - \vy,
\end{split}    
\end{align}
which indicates that all input samples invoke gradients which not only affect their own respective class weight vectors, but also others.

Thanks to these gradients, classifiers using cross-entropy loss with softmax can achieve a max-margin in decision.
However, these \textit{pull} and \textit{push} effects of softmax~\cite{softmax_pull_push} impede the maintenance of the classifier trained on previous tasks in continual learning.
We call the flow of gradients from a target class towards other classes as \textit{inter-task interference}.
General masked softmax can alleviate the inter-task interference by replacing the logits from input samples of old and new classes, which are not in a current task, to a specific masking value, \textit{i.e.},
\begin{align}        
\begin{split}
& \mathcal{L}_{\text{MSCE}} \left( \vx, \vy \right) 
= -\sum_{i=1}^{K} \evy_i \log ( \evp_i^{(t)} ),~~\text{where} \\
& \evp_i^{(t)} = 
\begin{cases}
\frac{\exp \left( \evz_i \right)}{
\sum_{j \in \sK^{(t)}} \exp(\evz_j) + (K - \lvert \sK^{(t)}\rvert) \cdot \exp(m)}, & \mbox{if }i \in \sK^{(t)} \\
\frac{\exp \left( m \right)}{
\sum_{j \in \sK^{(t)}} \exp(\evz_j) + (K - \lvert \sK^{(t)}\rvert) \cdot \exp(m)}, & \mbox{otherwise},
\end{cases}
\end{split}
\label{eq:msce}
\end{align}
where $\lvert \cdot \rvert$ denotes the cardinality of a set.

Eq.~\ref{eq:msce} means that class weight vectors except for those of the classes in the current task have a gradient of 0.
This stop gradient maintains the class weight vectors in previous and future tasks.
As a result, all gradients are calculated as below:
\begin{align}
    \begin{split}
        \frac{\partial \mathcal{L}_{\text{MSCE}}}{\partial z_k}
        &= p_k^{(t)} - 1, \\
        \frac{\partial \mathcal{L}_{\text{MSCE}}}{\partial z_i}
        &= p_k^{(t)},~~\forall i \in \sK_t \backslash \{k\}, \\
        \frac{\partial \mathcal{L}_{\text{MSCE}}}{\partial z_j}
        &= 0,~~\forall j \notin \sK_t. \\
    \end{split}
\end{align}

%
%

\begin{table*}[t]
\tiny
\centering
\caption{Final Average Accuracy $\uparrow$ (\%) for standard continual learning benchmarks. All experiments were conducted using 10 trials with random seeds except S-Tiny-ImageNet, where the experiment utilized 5 triasls with random seeds. Best in bold. (The performance notation: \textit{mean}$_{\textit{std}}$) (Please refer to the performance of commonly used methods in~\autoref{appx:tab:experiment_contid_learning_avg_acc} (\autoref{appx:sec:additional_experimental_results}))}
\begin{tabular}{clrrrrrrrr}
    \toprule
    \multirow{2}{*}{$\mathcal{B}$} & 
    \multirow{2}{*}{Method} &
    \multicolumn{2}{c}{S-MNIST} &
    \multicolumn{2}{c}{S-CIFAR-10} &
    \multicolumn{2}{c}{S-CIFAR-100} &
    \multicolumn{2}{c}{S-Tiny-ImageNet} \\
    & & 
    \textit{Class-IL} & \textit{Task-IL} &
    \textit{Class-IL} & \textit{Task-IL} &
    \textit{Class-IL} & \textit{Task-IL} &
    \textit{Class-IL} & \textit{Task-IL} \\
    \midrule
    \multirow{10}{*}{200} & ER{\scriptsize ~\cite{cil_replay_er}}
    & 80.43$_{1.89}$ & 97.86$_{0.35}$ & 44.79$_{1.86}$ & 91.19$_{0.94}$ 
    & \multicolumn{1}{c}{-} & \multicolumn{1}{c}{-} &  8.49$_{0.16}$ & 38.17$_{2.00}$ \\
    & DER{\scriptsize~\cite{cil_replay_derpp}}
    & 84.55$_{1.64}$ & 98.80$_{0.15}$ & 61.93$_{1.79}$ & 91.40$_{0.92}$ 
    & \multicolumn{1}{c}{-} & \multicolumn{1}{c}{-} & 11.87$_{0.78}$ & 40.22$_{0.67}$ \\
    & DER++{\scriptsize~\cite{cil_replay_derpp}}
    & 85.61$_{1.40}$ & 98.76$_{0.28}$ & 64.88$_{1.17}$ & 91.92$_{0.60}$ 
    & \multicolumn{1}{c}{-} & \multicolumn{1}{c}{-} & 10.96$_{1.17}$ & 40.87$_{1.16}$ \\
    \cmidrule(lr){2-10}
    & ER$^{\dag}$
    & 78.27$_{1.37}$ & 97.73$_{0.26}$ & 49.38$_{2.15}$ & 91.54$_{0.81}$ 
    & 14.88$_{0.47}$ & 66.75$_{1.06}$ &  8.58$_{0.19}$ & 38.39$_{0.72}$ \\
    & MSER($m=\text{-}\infty$)$^{\dag}$
    & 82.98$_{1.03}$ & 98.13$_{0.16}$ & 61.75$_{6.07}$ & 91.39$_{2.13}$
    & 28.51$_{0.44}$ & 68.51$_{0.87}$ & \textbf{15.47}$_{0.67}$ & 44.11$_{0.50}$ \\
    \cmidrule(lr){3-10}
    & DER++$^{\dag}$
    & 85.64$_{1.02}$ & 98.84$_{0.11}$ & 63.67$_{1.01}$ & 91.61$_{0.73}$ 
    & 24.85$_{1.69}$ & 67.69$_{1.39}$ & 11.59$_{1.07}$ & 41.00$_{0.88}$ \\
    & MSDER++($m=\text{-}\infty$)$^{\dag}$
    & 84.45$_{0.88}$ & 99.03$_{0.09}$ & \textbf{66.35}$_{1.52}$ & \textbf{93.17}$_{0.54}$ 
    & 28.57$_{1.11}$ & 74.02$_{0.76}$ & 13.21$_{0.56}$ & 49.75$_{0.99}$ \\
    & MSDER++($m=\text{-}1$)$^{\dag}$
    & \textbf{88.21}$_{0.49}$ & \textbf{99.07}$_{0.12}$ & 64.38$_{3.42}$ & 92.34$_{1.76}$
    & \textbf{28.70}$_{1.35}$ & \textbf{74.33}$_{0.72}$ & 13.25$_{0.51}$ & \textbf{51.24}$_{0.78}$ \\
    \midrule
    \multirow{10}{*}{500} & ER{\scriptsize~\cite{cil_replay_er}}
    & 86.12$_{1.89}$ & 99.04$_{0.18}$ & 57.74$_{0.27}$ & 93.61$_{0.27}$ 
    & \multicolumn{1}{c}{-} & \multicolumn{1}{c}{-} &  9.99$_{0.29}$ & 48.64$_{0.46}$ \\
    & DER{\scriptsize~\cite{cil_replay_derpp}}
    & 90.54$_{1.18}$ & 98.84$_{0.13}$ & 70.51$_{1.67}$ & 93.40$_{0.39}$ 
    & \multicolumn{1}{c}{-} & \multicolumn{1}{c}{-} & 17.75$_{1.14}$ & 51.78$_{0.88}$ \\
    & DER++{\scriptsize~\cite{cil_replay_derpp}}
    & 91.00$_{1.49}$ & 98.94$_{0.27}$ & 72.70$_{1.36}$ & 93.88$_{0.50}$ 
    & \multicolumn{1}{c}{-} & \multicolumn{1}{c}{-} & 19.38$_{1.41}$ & 51.91$_{0.68}$ \\
    \cmidrule(lr){2-10}
    & ER$^{\dag}$
    & 85.99$_{1.52}$ & 99.14$_{0.07}$ & 62.38$_{1.40}$ & 94.12$_{0.31}$ 
    & 21.53$_{0.69}$ & 73.97$_{0.30}$ & 10.12$_{0.22}$ & 48.06$_{0.80}$ \\
    & MSER($m=\text{-}\infty$)$^{\dag}$
    & 89.35$_{0.59}$ & \textbf{99.20}$_{0.16}$ & 70.64$_{1.28}$ & 94.22$_{0.41}$
    & 35.68$_{0.89}$ & 74.77$_{0.71}$ & \textbf{20.43}$_{0.38}$ & 53.21$_{0.84}$ \\
    \cmidrule(lr){3-10}
    & DER++$^{\dag}$
    & \textbf{91.01}$_{0.46}$ & 98.95$_{0.07}$ & 73.15$_{0.80}$ & 94.07$_{0.39}$ 
    & 37.41$_{1.40}$ & 76.07$_{0.60}$ & 19.82$_{0.87}$ & 52.24$_{0.94}$ \\
    & MSDER++($m=\text{-}\infty$)$^{\dag}$
    & 83.10$_{1.22}$ & 99.08$_{0.09}$ & 71.85$_{3.76}$ & 94.28$_{1.49}$ 
    & 37.80$_{0.92}$ & 80.52$_{0.60}$ & 17.71$_{0.58}$ & 59.86$_{1.08}$ \\
    & MSDER++($m=\text{-}1$)$^{\dag}$
    & 88.41$_{1.05}$ & 99.06$_{0.07}$ & \textbf{73.53}$_{0.78}$ & \textbf{94.52}$_{0.47}$
    & \textbf{38.76}$_{1.23}$ & \textbf{80.96}$_{0.41}$ & 17.68$_{0.54}$ & \textbf{60.85}$_{0.91}$ \\
    \midrule
    \multirow{10}{*}{5120} & ER{\scriptsize~\cite{cil_replay_er}}
    & 93.40$_{1.29}$ & 99.33$_{0.22}$ & 82.47$_{0.52}$ & 96.98$_{0.17}$ 
    & \multicolumn{1}{c}{-} & \multicolumn{1}{c}{-} & 27.40$_{0.31}$ & 67.29$_{0.23}$ \\
    & DER{\scriptsize~\cite{cil_replay_derpp}}
    & 94.90$_{0.57}$ & 99.29$_{0.11}$ & 83.81$_{0.33}$ & 95.43$_{0.33}$ 
    & \multicolumn{1}{c}{-} & \multicolumn{1}{c}{-} & 36.73$_{0.64}$ & 69.50$_{0.26}$ \\
    & DER++{\scriptsize~\cite{cil_replay_derpp}}
    & 95.30$_{1.20}$ & 99.47$_{0.07}$ & 85.24$_{0.49}$ & 96.12$_{0.21}$
    & \multicolumn{1}{c}{-} & \multicolumn{1}{c}{-} & 39.02$_{0.97}$ & 69.84$_{0.63}$ \\
    \cmidrule(lr){2-10}
    & ER$^{\dag}$
    & 93.42$_{1.08}$ & 99.41$_{0.15}$ & 84.31$_{0.38}$ & \textbf{97.02}$_{0.26}$ 
    & 50.51$_{0.94}$ & 85.53$_{0.73}$ & 27.30$_{0.51}$ & 67.69$_{0.33}$ \\
    & MSER($m=\text{-}\infty$)$^{\dag}$
    & 93.51$_{0.60}$ & 99.38$_{0.12}$ & 82.63$_{1.34}$ & 96.45$_{0.27}$
    & 52.95$_{0.73}$ & 84.20$_{0.58}$ & 35.73$_{0.41}$ & 67.50$_{0.53}$ \\
    \cmidrule(lr){3-10}
    & DER++$^{\dag}$
    & \textbf{95.09}$_{0.56}$ & 99.50$_{0.08}$ & 85.56$_{0.38}$ & 96.30$_{0.22}$ 
    & \textbf{59.62}$_{0.63}$ & 86.61$_{0.32}$ & \textbf{39.66}$_{0.89}$ & 69.95$_{0.32}$ \\
    & MSDER++($m=\text{-}\infty$)$^{\dag}$
    & 93.75$_{0.23}$ & \textbf{99.62}$_{0.05}$ & 84.71$_{0.65}$ & 96.78$_{0.16}$ 
    & 58.18$_{0.43}$ & 87.97$_{0.33}$ & 34.72$_{0.46}$ & 72.40$_{0.25}$ \\
    & MSDER++($m=\text{-}1$)$^{\dag}$
    & 94.36$_{0.22}$ & 99.57$_{0.07}$ & \textbf{85.66}$_{0.54}$ & 96.91$_{0.16}$
    & 59.09$_{0.38}$ & \textbf{88.41}$_{0.22}$ & 35.30$_{0.31}$ & \textbf{73.11}$_{0.12}$ \\
    \bottomrule
\end{tabular}
\label{tab:experiment_contid_learning_avg_acc}
\end{table*}

\begin{table*}[t]
\tiny
\centering
\caption{Final Average Forget $\downarrow$ (\%) for standard continual learning benchmarks. All experiments were conducted using 10 trials with random seeds except S-Tiny-ImageNet, where the experiment utilized 5 trials with random seeds. Best in bold. (The performance notation: \textit{mean}$_{\textit{std}}$) (Please refer to the performance of commonly used methods in~\autoref{appx:tab:experiment_contid_learning_avg_forget} (\autoref{appx:sec:additional_experimental_results}))}
\begin{tabular}{clrrrrrrrr}
    \toprule
    \multirow{2}{*}{$\mathcal{B}$} & 
    \multirow{2}{*}{Method} &
    \multicolumn{2}{c}{S-MNIST} &
    \multicolumn{2}{c}{S-CIFAR-10} &
    \multicolumn{2}{c}{S-CIFAR-100} &
    \multicolumn{2}{c}{S-Tiny-ImageNet} \\
    & & 
    \textit{Class-IL} & \textit{Task-IL} &
    \textit{Class-IL} & \textit{Task-IL} &
    \textit{Class-IL} & \textit{Task-IL} &
    \textit{Class-IL} & \textit{Task-IL} \\
    \midrule
    \multirow{10}{*}{200} & ER{\scriptsize~\cite{cil_replay_er}}
    & 21.36$_{2.46}$ &  0.84$_{0.41}$ & 61.24$_{2.62}$ &  7.08$_{0.64}$
    & \multicolumn{1}{c}{-} & \multicolumn{1}{c}{-} 
    & \multicolumn{1}{c}{-} & \multicolumn{1}{c}{-} \\
    & DER{\scriptsize~\cite{cil_replay_derpp}}
    & 17.66$_{2.10}$ &  0.57$_{0.18}$ & 40.76$_{0.42}$ &  6.57$_{0.20}$
    & \multicolumn{1}{c}{-} & \multicolumn{1}{c}{-} 
    & \multicolumn{1}{c}{-} & \multicolumn{1}{c}{-} \\
    & DER++{\scriptsize~\cite{cil_replay_derpp}}
    & 16.27$_{1.73}$ &  0.66$_{0.28}$ & 32.59$_{2.32}$ &  5.16$_{0.21}$
    & \multicolumn{1}{c}{-} & \multicolumn{1}{c}{-} 
    & \multicolumn{1}{c}{-} & \multicolumn{1}{c}{-} \\
    \cmidrule(lr){2-10}
    & ER$^{\dag}$
    & 24.09$_{1.61}$ &  0.89$_{0.24}$ & 58.97$_{2.70}$ &  6.49$_{0.90}$ 
    & 81.61$_{0.61}$ & 24.18$_{1.12}$ & 76.62$_{0.48}$ & 43.52$_{0.85}$ \\
    & MSER($m=\text{-}\infty$)$^{\dag}$
    & 10.22$_{1.20}$ &  0.69$_{0.12}$ & 26.30$_{6.18}$ &  6.18$_{1.56}$
    & \textbf{46.51}$_{1.36}$ & 20.92$_{0.79}$ & \textbf{50.86}$_{0.79}$ & 34.97$_{0.61}$ \\
    \cmidrule(lr){3-10}
    & DER++$^{\dag}$
    & 16.34$_{1.23}$ &  0.53$_{0.12}$ & 34.80$_{1.68}$ &  6.45$_{1.07}$ 
    & 68.80$_{2.76}$ & 24.20$_{1.26}$ & 72.77$_{2.13}$ & 41.11$_{0.97}$ \\
    & MSDER++($m=\text{-}\infty$)$^{\dag}$
    &  \textbf{4.29}$_{0.75}$ &  \textbf{0.32}$_{0.09}$ & 23.10$_{1.46}$ &  \textbf{4.82}$_{0.71}$ 
    & 48.52$_{0.62}$ & 17.34$_{0.77}$ & 52.96$_{0.72}$ & 32.20$_{1.21}$ \\
    & MSDER++($m=\text{-}1$)$^{\dag}$
    &  5.00$_{0.51}$ &  \textbf{0.32}$_{0.10}$ & \textbf{22.84}$_{2.00}$ &  5.50$_{1.47}$
    & 48.48$_{1.59}$ & \textbf{17.09}$_{0.77}$ & 55.22$_{0.56}$ & \textbf{30.32}$_{0.93}$ \\
    \midrule
    \multirow{10}{*}{500} & ER{\scriptsize~\cite{cil_replay_er}}
    & 15.97$_{2.46}$ &  0.39$_{0.20}$ & 45.35$_{0.07}$ &  3.54$_{0.35}$
    & \multicolumn{1}{c}{-} & \multicolumn{1}{c}{-} 
    & \multicolumn{1}{c}{-} & \multicolumn{1}{c}{-} \\
    & DER{\scriptsize~\cite{cil_replay_derpp}}
    &  9.58$_{1.52}$ &  0.45$_{0.13}$ & 26.74$_{0.15}$ &  4.56$_{0.45}$
    & \multicolumn{1}{c}{-} & \multicolumn{1}{c}{-} 
    & \multicolumn{1}{c}{-} & \multicolumn{1}{c}{-} \\
    & DER++{\scriptsize~\cite{cil_replay_derpp}}
    &  8.85$_{1.86}$ &  0.35$_{0.15}$ & 22.38$_{4.41}$ &  4.66$_{1.15}$
    & \multicolumn{1}{c}{-} & \multicolumn{1}{c}{-} 
    & \multicolumn{1}{c}{-} & \multicolumn{1}{c}{-} \\
    \cmidrule(lr){2-10}
    & ER$^{\dag}$
    & 16.28$_{1.99}$ &  0.38$_{0.11}$ & 42.83$_{1.89}$ &  3.51$_{0.40}$ 
    & 73.90$_{0.82}$ & 16.17$_{0.40}$ & 74.82$_{0.08}$ & 32.17$_{1.01}$ \\
    & MSER($m=\text{-}\infty$)$^{\dag}$
    &  7.18$_{1.37}$ &  0.39$_{0.17}$ & 17.31$_{1.58}$ &  3.09$_{0.36}$
    & 39.74$_{1.36}$ & 13.94$_{0.64}$ & 51.25$_{0.35}$ & 27.73$_{0.80}$ \\
    \cmidrule(lr){3-10}
    & DER++$^{\dag}$
    &  9.00$_{0.50}$ &  0.41$_{0.05}$ & 23.12$_{1.88}$ &  3.56$_{0.38}$ 
    & 52.74$_{2.65}$ & 15.13$_{0.74}$ & 59.04$_{1.27}$ & 27.96$_{0.72}$ \\
    & MSDER++($m=\text{-}\infty$)$^{\dag}$
    &  \textbf{2.07}$_{0.89}$ &  \textbf{0.23}$_{0.09}$ & 16.95$_{3.77}$ &  \textbf{3.03}$_{0.76}$ 
    & 35.98$_{0.79}$ & 10.39$_{0.68}$ & \textbf{47.00}$_{0.75}$ & 21.53$_{1.00}$ \\
    & MSDER++($m=\text{-}1$)$^{\dag}$
    &  4.00$_{0.78}$ &  0.27$_{0.07}$ & \textbf{14.70}$_{1.31}$ &  3.21$_{0.61}$
    & \textbf{35.82}$_{0.99}$ & \textbf{10.00}$_{0.38}$ & 47.25$_{1.04}$ & \textbf{20.20}$_{1.40}$ \\
    \midrule
    \multirow{10}{*}{5120} & ER{\scriptsize~\cite{cil_replay_er}}
    &  6.08$_{1.84}$ &  0.25$_{0.23}$ & 13.99$_{0.12}$ &  0.27$_{0.06}$
    & \multicolumn{1}{c}{-} & \multicolumn{1}{c}{-} 
    & \multicolumn{1}{c}{-} & \multicolumn{1}{c}{-} \\
    & DER{\scriptsize~\cite{cil_replay_derpp}}
    &  4.53$_{0.83}$ &  0.32$_{0.08}$ & 10.12$_{0.80}$ &  2.59$_{0.08}$
    & \multicolumn{1}{c}{-} & \multicolumn{1}{c}{-} 
    & \multicolumn{1}{c}{-} & \multicolumn{1}{c}{-} \\
    & DER++{\scriptsize~\cite{cil_replay_derpp}}
    &  4.19$_{1.63}$ &  0.23$_{0.06}$ &  7.27$_{0.84}$ &  1.18$_{0.19}$
    & \multicolumn{1}{c}{-} & \multicolumn{1}{c}{-} 
    & \multicolumn{1}{c}{-} & \multicolumn{1}{c}{-} \\
    \cmidrule(lr){2-10}
    & ER$^{\dag}$
    &  6.26$_{1.56}$ &  0.21$_{0.10}$ & 14.44$_{0.65}$ &  \textbf{0.46}$_{0.21}$ 
    & 39.25$_{0.62}$ &  4.53$_{0.36}$ & 54.54$_{0.47}$ & 11.60$_{0.25}$ \\
    & MSER($m=\text{-}\infty$)$^{\dag}$
    &  3.12$_{1.10}$ &  0.24$_{0.10}$ &  7.57$_{1.31}$ &  0.81$_{0.28}$
    & 22.01$_{0.40}$ &  4.10$_{0.35}$ & 30.03$_{0.58}$ &  9.50$_{0.52}$ \\
    \cmidrule(lr){3-10}
    & DER++$^{\dag}$
    &  4.59$_{0.79}$ &  0.24$_{0.10}$ &  7.50$_{0.80}$ &  1.05$_{0.34}$ 
    & 25.28$_{0.71}$ &  4.56$_{0.50}$ & 31.60$_{1.79}$ &  9.63$_{0.53}$ \\
    & MSDER++($m=\text{-}\infty$)$^{\dag}$
    &  \textbf{0.90}$_{0.18}$ &  \textbf{0.09}$_{0.06}$ &  \textbf{5.02}$_{0.69}$ &  0.71$_{0.27}$ 
    & \textbf{14.79}$_{0.83}$ &  2.90$_{0.44}$ & \textbf{20.30}$_{0.48}$ &  7.23$_{0.20}$ \\
    & MSDER++($m=\text{-}1$)$^{\dag}$
    &  1.20$_{0.31}$ &  0.13$_{0.05}$ &  5.43$_{0.56}$ &  0.51$_{0.14}$
    & 15.34$_{0.52}$ &  \textbf{2.60}$_{0.27}$ & 21.39$_{0.74}$ &  \textbf{6.86}$_{0.35}$ \\
    \bottomrule
\end{tabular}
\label{tab:experiment_contid_learning_avg_forget}
\end{table*}

\section{Experimental Results}
\paragraph{Common Settings.}
We calculated the mean and standard deviation of experimental results in multiple times using different random seeds.
The number of seeds used in each experiment will be described in the corresponding table.
$\dag$ means that the experiments was conducted on the same environmental setting to ours.

\paragraph{Evaluation Metrics.}
Two commonly used evaluation metrics for quantitatively assessing model performance in continual learning settings are Final Average Accuracy $(A_T)$ and Final Average Forgetting $(F_T)$. 
Following~\cite{contid_metric}, we denote them as Eq.~\ref{eq:metric}.
\begin{align}
\begin{split}
    A_T &= \frac{1}{T} \sum_{t=1}^{T} a_{T,t}, \\
    F_T &= \frac{1}{T-1} \sum_{t=1}^{T} \max_{i \in \{1, \dots T-1\}} (a_{i,t} - a_{T,t}),
\end{split}    
\label{eq:metric}
\end{align}
where $a_{j,t}$ denotes the test accuracy on the task $T_t$ after the model has been trained on all tasks up to the task $T_j$ . 
A higher test accuracy on each task after training is indicative of a higher $A_T$, while a lower reduction of test accuracy on each task after training is indicative of a lower $F_T$.

\paragraph{Datasets.}
We followed~\cite{cil_replay_derpp} and conducted comparison experiments on a split datasets of the MNIST, CIFAR-10, CIFAR100, and Tiny-ImageNet.
The split-MNIST dataset consists of 5 tasks with 2 classes sequencially divided to each task, \textit{i.e.}, the task $T_0$ has 0 and 1 digits and the last task $T_4$ has 8 and 9 digits.
In the same way, the split-CIFAR10 dataset is composed to 5 tasks with 2 classes.
For the split-CIFAR100 and split-Tiny-ImageNet, we set 10 tasks with 10 classes and 20 tasks with 10 classes, respectively.

\paragraph{Architectures.}
For comparison experiments with~\cite{cil_replay_derpp} in equivalent conditions, we follow~\cite{cil_replay_derpp} and conducted the experiments by employing a fully-connected network with two hidden layers, each one comprising of 100 ReLU units for the split-MNIST.
We also use other split datasets with ResNet18~\cite{arch_resnet}.

\begin{table}
    \caption{Hyperparameter settings on all experiments in continual learning scenarios. ($\mathcal{B}$: buffer size, $lr$: learning rate, $\alpha$: weight hyperparameter for the cross-entropy loss from buffer samples, $\beta$: weight hyperparameter for distillation loss from buffer samples.)}
    \scriptsize
    \centering
    \begin{tabular}{|c|c|c|c|c|c|}
        \hline
        \textbf{Method} & \textbf{$\mathcal{B}$} & \textbf{S-MNIST} & \textbf{S-CIFAR10} & \textbf{S-CIFAR100} & \textbf{S-TinyImg} \\
        \hline
        \multirow{3}{*}{ER} 
        & 200 
        & \textit{lr}: 0.01
        & \textit{lr}: 0.1
        & \textit{lr}: 0.1
        & \textit{lr}: 0.1
        \\
        & 500
        & \textit{lr}: 0.1
        & \textit{lr}: 0.1
        & \textit{lr}: 0.1
        & \textit{lr}: 0.03
        \\
        & 5120
        & \textit{lr}: 0.1
        & \textit{lr}: 0.1
        & \textit{lr}: 0.1
        & \textit{lr}: 0.1
        \\
        \hline
        \multirow{9}{*}{DER++} 
        & \multirow{3}{*}{200}
        & \textit{lr}: 0.03, 
        & \textit{lr}: 0.03, 
        & \textit{lr}: 0.03, 
        & \textit{lr}: 0.03, \\
        & 
        & $\alpha$: 0.2, 
        & $\alpha$: 0.1, 
        & $\alpha$: 0.1, 
        & $\alpha$: 0.1, \\
        & 
        & $\beta$: 1.0 
        & $\beta$: 0.5 
        & $\beta$: 0.5 
        & $\beta$: 1.0 
        \\
        \cline{2-6}
        & \multirow{3}{*}{500}
        & \textit{lr}: 0.03,
        & \textit{lr}: 0.03,
        & \textit{lr}: 0.03,
        & \textit{lr}: 0.03, \\
        &
        & $\alpha$: 1.0,
        & $\alpha$: 0.2,
        & $\alpha$: 0.1,
        & $\alpha$: 0.2, \\
        &
        & $\beta$: 0.5 
        & $\beta$: 0.5 
        & $\beta$: 0.5 
        & $\beta$: 0.5
        \\
        \cline{2-6}
        & \multirow{3}{*}{5120}
        & \textit{lr}: 0.1,
        & \textit{lr}: 0.03,
        & \textit{lr}: 0.03,
        & \textit{lr}: 0.03, \\
        &
        & $\alpha$: 0.2,
        & $\alpha$: 0.1,
        & $\alpha$: 0.1,
        & $\alpha$: 0.1, \\
        &
        & $\beta$: 0.5 
        & $\beta$: 1.0
        & $\beta$: 0.5 
        & $\beta$: 0.5 
        \\
        \hline
    \end{tabular}
    \label{tab:hyperparameter_settings}
\end{table}

\subsection{Continual Learning Benchmarks}
\label{subsec:cl_benchmarks}

\paragraph{Implementation Details.}
We followed the best hyperparameter settings from~\cite{cil_replay_derpp}, where comprehensive hyperparameter search was conducted.
We summarize them in~\autoref{tab:hyperparameter_settings} and our method uses the same hyperparameter settings to each method.
Batch sizes are fixed to each dataset: 10 in the split MNIST dataset, 32 in split CIFAR-10, CIFAR-100, and Tiny-ImageNet datasets.

\paragraph{Results and Analyses.}
As illustrated in~\autoref{tab:experiment_contid_learning_avg_acc} and~\autoref{tab:experiment_contid_learning_avg_forget}, our method has outperformed in both class-incremental learning and task-incremental learning. 
In class-incremental learning, our approach also showed superior performance in most cases, particularly in scenarios with low buffer sizes and more complex datasets (refer to buffer size set to 200 in S-CIFAR100 and S-Tiny-ImageNet in~\autoref{tab:experiment_contid_learning_avg_acc}). 
Even in task-incremental learning, the masked replay method consistently outperformed in all experiments. 
This indicates that the masked softmax effectively maintains previously trained information by alleviating inter-task interference.

\paragraph{Distillation with Masked Softmax is Dangerous.}
As shown in~\autoref{tab:experiment_contid_learning_avg_acc} and~\autoref{tab:experiment_contid_learning_avg_forget}, the use of masked softmax with negative infinity significantly improves model stability, especially when used with ER. 
However, when used with knowledge distillation, the performance of most cases has decreased compared to the baseline, even lower than MSER, where there is no danger due to distillation (refer to MSDER++($m$=-$\infty$)).

\paragraph{Control Model Stability by Masking Value Adjustment.}
As shown in~\autoref{tab:experiment_contid_learning_avg_acc}, in particular, this danger stands out in simple and low buffer size settings. 
However, when the masking value is set to -1, the performance of most models improves. 
Therefore, we conclude that controlling the masking value is helpful in mitigating the extreme increase in model stability.

\begin{table}[t]
\tiny
\centering
\caption{Classification results for extremely low buffer sizes in standard continual learning benchmarks. All experiments were conducted using 10 trials with random seeds. $(\cdot)$ means the masking value used in each experiment. Best in bold. ($A_T$: Final Average Accuracy $\uparrow$ (\%), $F_T$: Final Average Forgetting $\downarrow$ (\%))} 
\begin{tabular}{clrrrrrrrr}
    \toprule
    \multirow{4}{*}{$\mathcal{B}$} & 
    \multirow{4}{*}{Method} &
    \multicolumn{4}{c}{S-MNIST} &
    \multicolumn{4}{c}{S-CIFAR-10} \\
    & & 
    \multicolumn{2}{c}{\textit{Class-IL}} & 
    \multicolumn{2}{c}{\textit{Task-IL}} &
    \multicolumn{2}{c}{\textit{Class-IL}} & 
    \multicolumn{2}{c}{\textit{Task-IL}} \\
    \cmidrule(rl){3-4} \cmidrule(rl){5-6} \cmidrule(rl){7-8} \cmidrule(rl){9-10}
    & &
    \multicolumn{1}{c}{$A_T$} & \multicolumn{1}{c}{$F_T$} &
    \multicolumn{1}{c}{$A_T$} & \multicolumn{1}{c}{$F_T$} &
    \multicolumn{1}{c}{$A_T$} & \multicolumn{1}{c}{$F_T$} &
    \multicolumn{1}{c}{$A_T$} & \multicolumn{1}{c}{$F_T$} \\
    \midrule
    \multirow{3}{*}{10} 
    & DER++$^{\dag}$
    & 32.91$_{3.45}$ & 82.79$_{4.30}$
    & 95.12$_{4.20}$ &  5.08$_{5.26}$ 
    & 28.50$_{4.05}$ & 71.00$_{8.62}$
    & 71.31$_{3.60}$ & 31.26$_{4.77}$ \\
    & MSDER++($\text{-}\infty$)$^{\dag}$
    & \textbf{58.80}$_{4.47}$ & \textbf{10.27}$_{3.45}$
    & \textbf{98.82}$_{0.12}$ &  \textbf{0.59}$_{0.14}$
    & \textbf{30.33}$_{2.59}$ & 48.06$_{12.00}$
    & \textbf{73.76}$_{6.53}$ & \textbf{28.71}$_{8.37}$ \\
    & MSDER++($\text{-}1$)$^{\dag}$
    & 53.29$_{4.12}$ & 17.59$_{5.85}$
    & 96.95$_{1.94}$ &  2.83$_{2.39}$
    & 20.16$_{2.62}$ & \textbf{32.77}$_{7.67}$
    & 72.15$_{4.65}$ & 29.90$_{5.72}$ \\
    \midrule
    \multirow{3}{*}{50} 
    & DER++$^{\dag}$
    & 63.66$_{2.58}$ & 44.27$_{3.21}$
    & 98.26$_{0.42}$ &  1.29$_{0.51}$
    & 47.08$_{2.79}$ & 54.57$_{3.40}$
    & 85.29$_{2.46}$ & 14.15$_{3.33}$ \\
    & MSDER++($\text{-}\infty$)$^{\dag}$
    & 76.81$_{2.43}$ &  \textbf{8.49}$_{3.19}$
    & \textbf{99.04}$_{0.10}$ &  \textbf{0.33}$_{0.07}$
    & \textbf{49.79}$_{7.19}$ & 37.49$_{4.97}$
    & \textbf{86.82}$_{5.96}$ & \textbf{11.96}$_{6.70}$ \\
    & MSDER++($\text{-}1$)$^{\dag}$
    & \textbf{79.07}$_{2.63}$ & 10.34$_{2.70}$
    & 98.79$_{0.19}$ &  0.62$_{0.21}$
    & 41.21$_{5.81}$ & \textbf{29.40}$_{4.17}$
    & 82.65$_{3.29}$ & 17.75$_{4.01}$ \\
    \midrule
    \multirow{3}{*}{100} 
    & DER++$^{\dag}$
    & 76.31$_{2.50}$ & 28.21$_{3.14}$
    & 98.57$_{0.17}$ &  0.86$_{0.24}$
    & 55.25$_{1.33}$ & 45.82$_{2.35}$
    & 88.88$_{1.03}$ &  9.97$_{1.32}$ \\
    & MSDER++($\text{-}\infty$)$^{\dag}$
    & 81.06$_{1.46}$ &  \textbf{4.83}$_{1.30}$
    & \textbf{99.02}$_{0.06}$ &  \textbf{0.35}$_{0.10}$
    & \textbf{58.95}$_{1.53}$ & 30.86$_{1.83}$
    & \textbf{91.33}$_{0.89}$ &  \textbf{6.52}$_{1.21}$ \\
    & MSDER++($\text{-}1$)$^{\dag}$
    & \textbf{84.16}$_{1.90}$ &  6.83$_{1.02}$
    & 98.91$_{0.14}$ &  0.46$_{0.19}$
    & 53.90$_{1.92}$ & \textbf{28.14}$_{2.78}$
    & 88.49$_{2.80}$ & 10.21$_{3.51}$ \\
    \bottomrule
\end{tabular}
\label{tab:experiment_low_buffer_avg_acc}
\end{table}

\subsection{Continual Learning in Low Buffer Size}
\label{subsec:cl_lowbuf}

\paragraph{Implementation Details.}
Based on the findings from the results of softmax masking experiments on continual learning benchmarks, it was found that softmax masking is more effective when the buffer size is lower.
We demonstrated this by conducting various experiments with extremely constrained buffer sizes on the split MNIST and CIFAR10 datasets. 
The episodic memory buffer size was reduced to 10, 50, and 100, and the model was trained with and without masked softmax. 
The results were then quantitatively compared. 
All experimental settings, except for the buffer size, are equivalent to those used in continual learning benchmarks.

\paragraph{Results and Analyses.}
As shown in~\autoref{tab:experiment_low_buffer_avg_acc}, MSDER++(-1) outperforms DER++ and MSDER++(-$\infty$) in 10 and 50 buffer size on the split MNIST dataset.
However, MSDER++(-$\infty$) consistently shows its superior performance in other experiments. 
This suggests that when the replay samples are insufficient to recover the previous knowledge, increasing the model's stability is a more effective method for improving performance in continual learning scenarios than transferring knowledge from previous tasks.

\begin{figure}[t]
\centering
\begin{subfigure}[t]{0.32\linewidth}
    \centering
    \includegraphics[width=\linewidth]{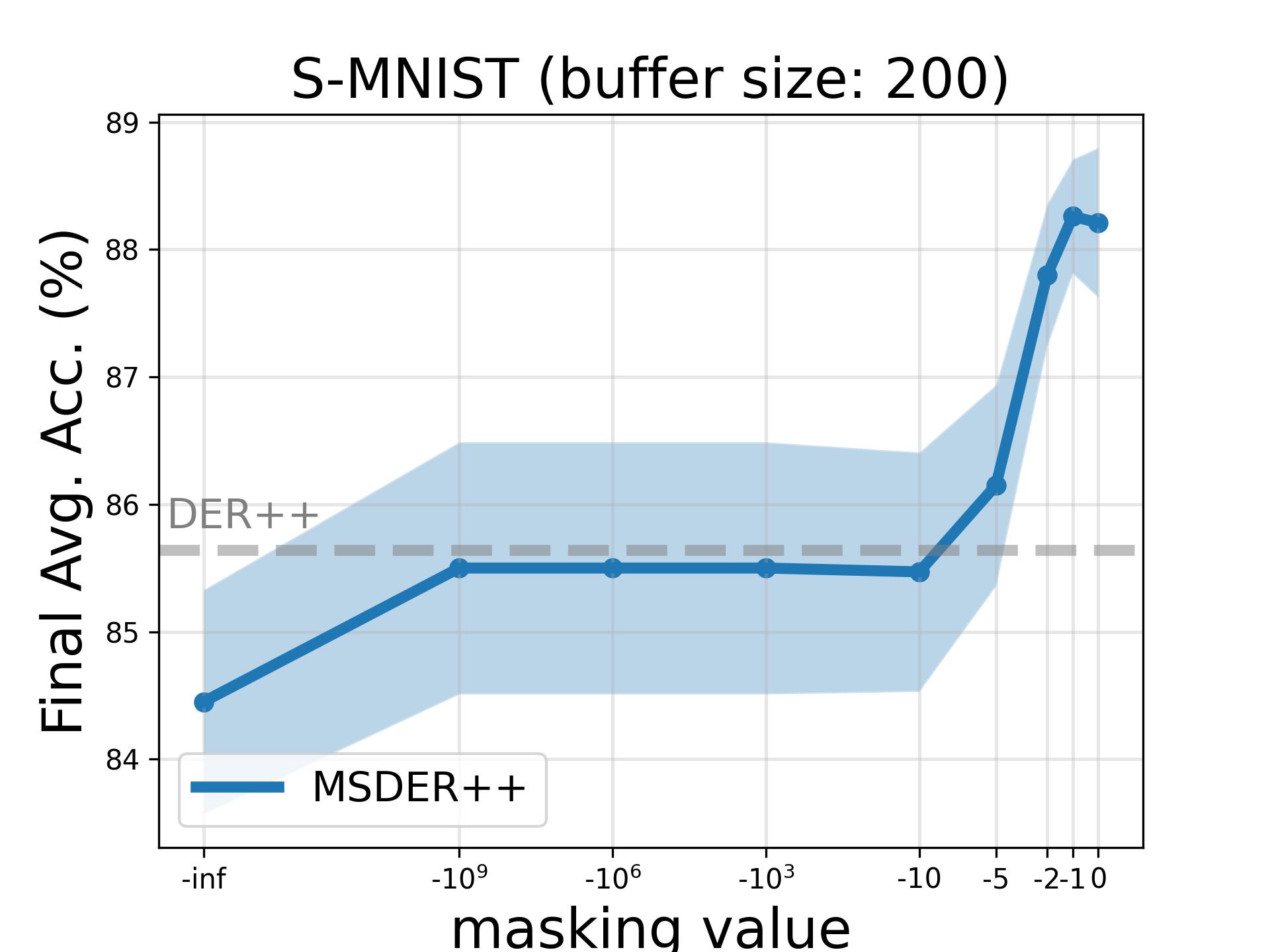}
    \caption{S-MNIST ($\mathcal{B}: 200$)}
    \label{fig:analysis:seq-mnist_buf200}
\end{subfigure}
\begin{subfigure}[t]{0.32\linewidth}
    \centering
    \includegraphics[width=\linewidth]{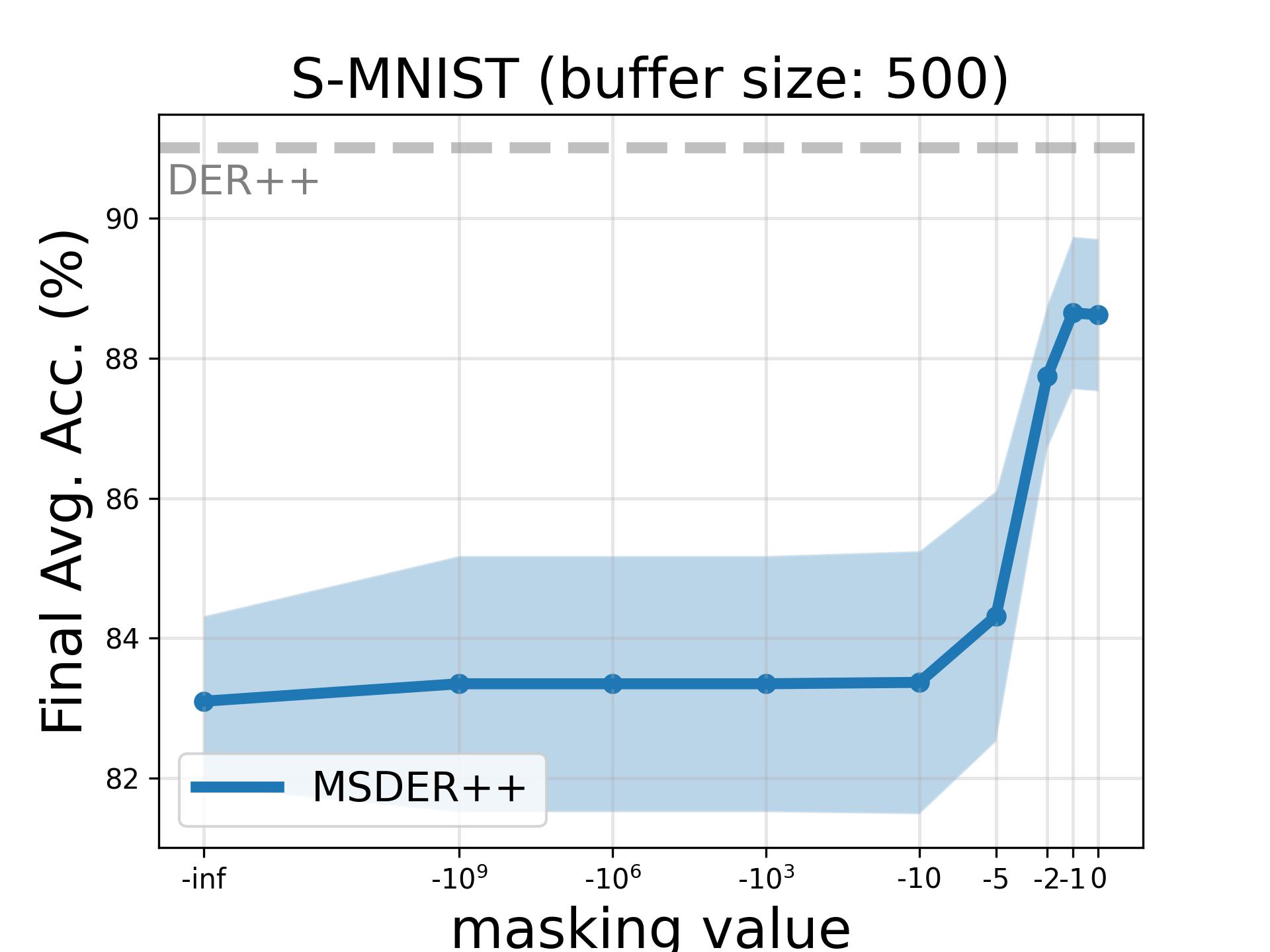}
    \caption{S-MNIST ($\mathcal{B}: 500$)}
    \label{fig:analysis:seq-mnist_buf500}
\end{subfigure}
\begin{subfigure}[t]{0.32\linewidth}
    \centering
    \includegraphics[width=\linewidth]{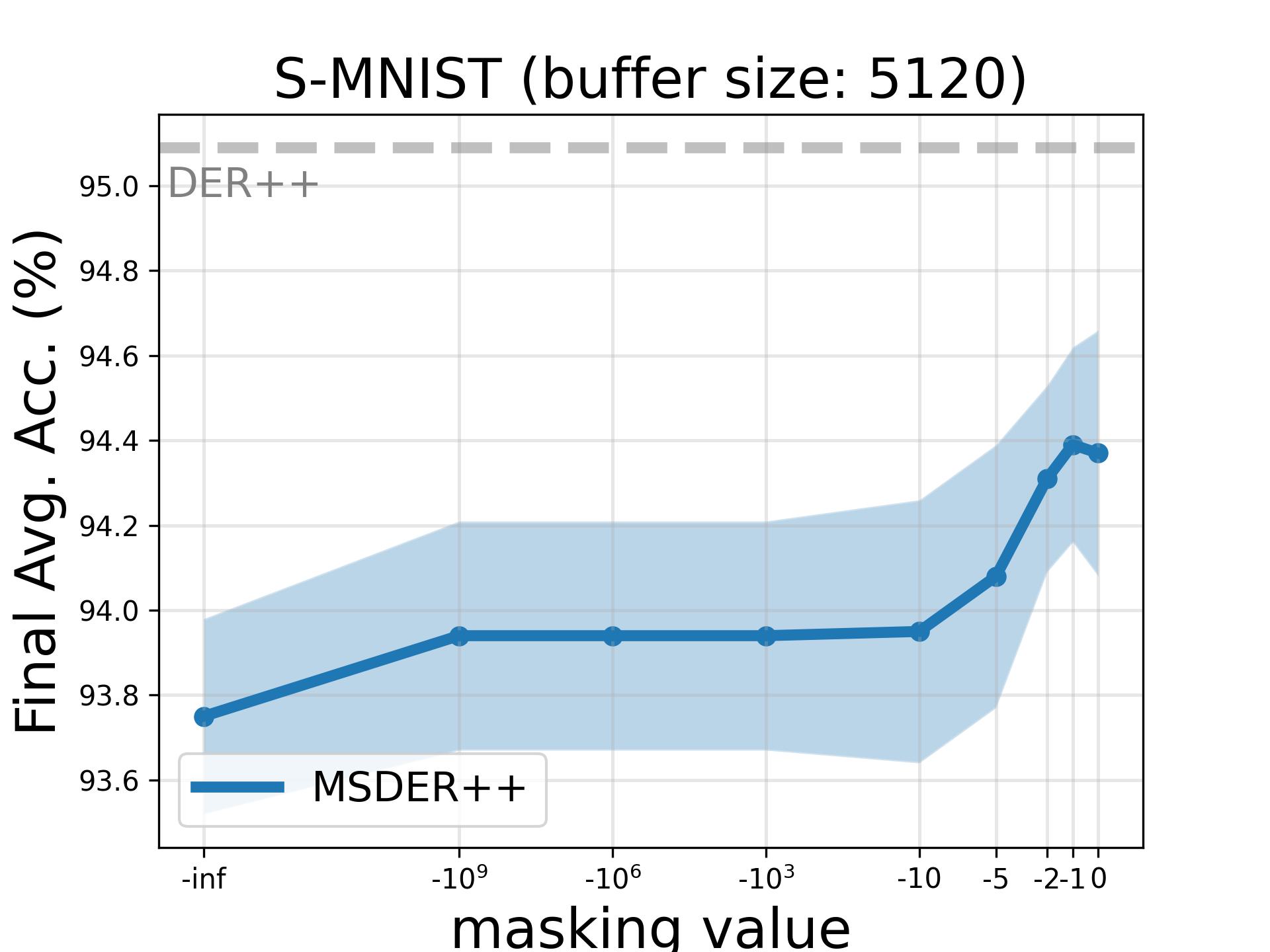}
    \caption{S-MNIST ($\mathcal{B}: 5120$)}
    \label{fig:analysis:seq-mnist_buf5120}
\end{subfigure}
\begin{subfigure}[t]{0.32\linewidth}
    \centering
    \includegraphics[width=\linewidth]{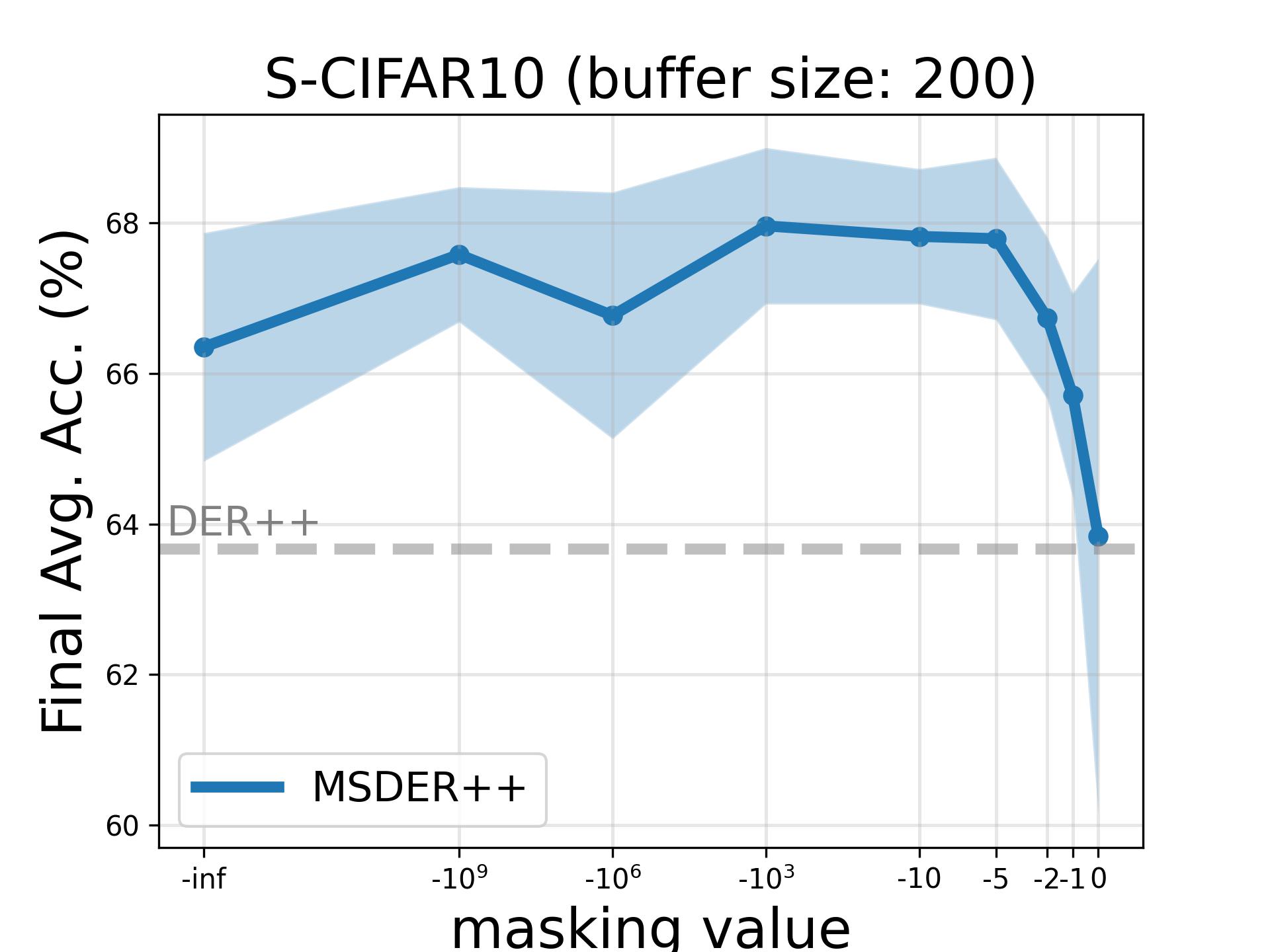}
    \caption{S-CIFAR10 ($\mathcal{B}: 200$)}
    \label{fig:analysis:seq-cifar10_buf200}
\end{subfigure}
\begin{subfigure}[t]{0.32\linewidth}
    \centering
    \includegraphics[width=\linewidth]{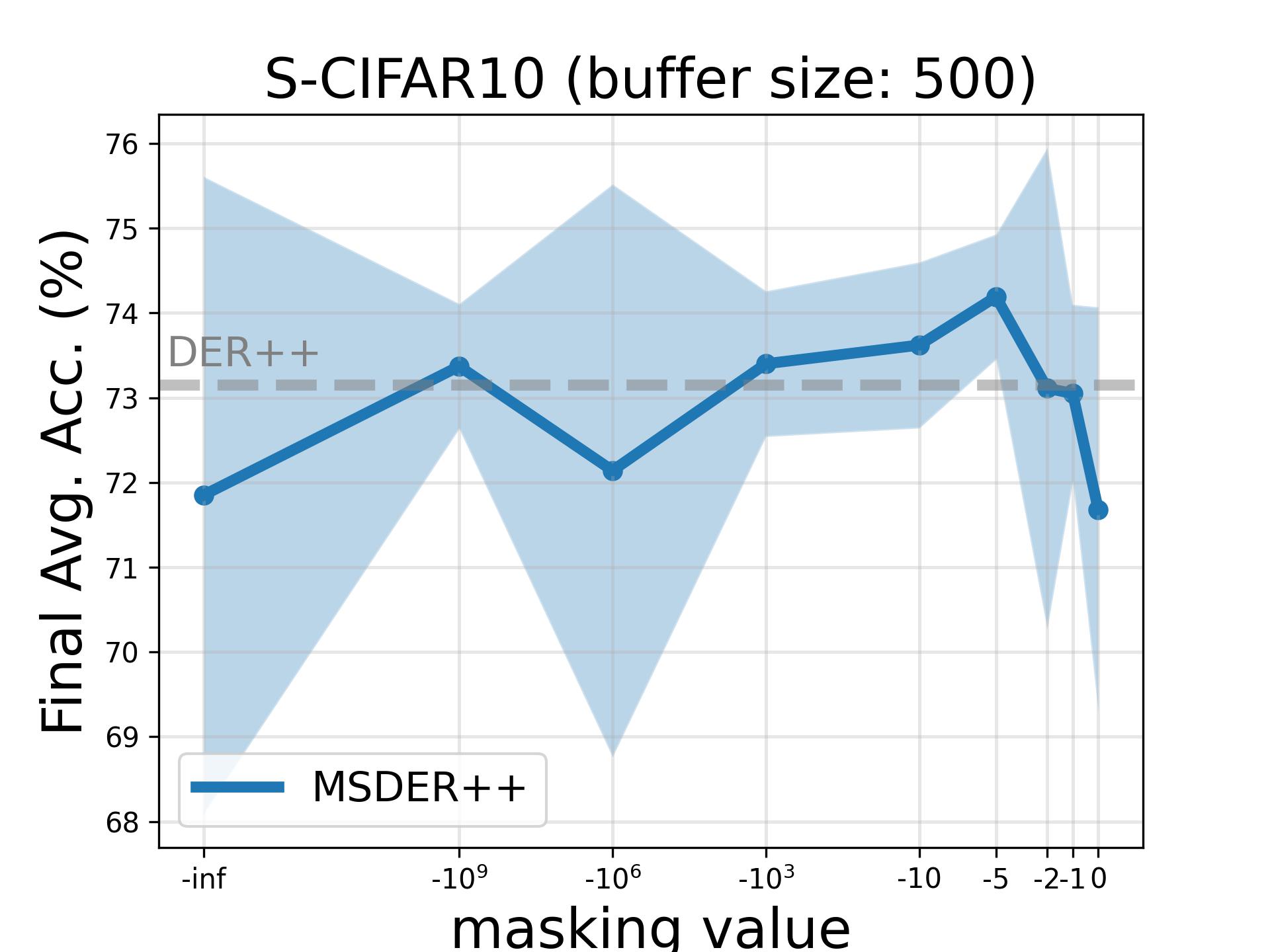}
    \caption{S-CIFAR10 ($\mathcal{B}: 500$)}
    \label{fig:analysis:seq-cifar10_buf500}
\end{subfigure}
\begin{subfigure}[t]{0.32\linewidth}
    \centering
    \includegraphics[width=\linewidth]{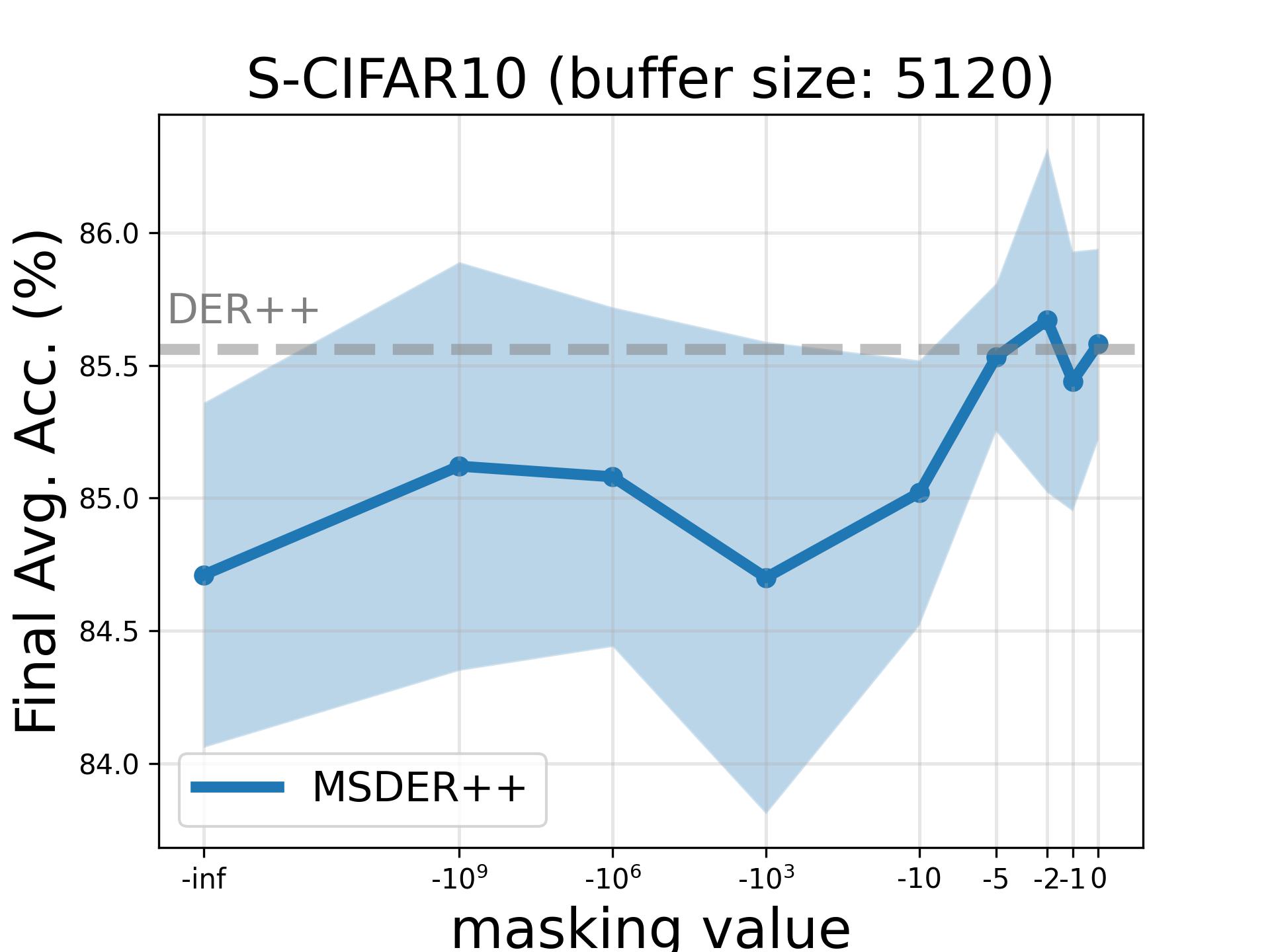}
    \caption{S-CIFAR10 ($\mathcal{B}: 5120$)}
    \label{fig:analysis:seq-cifar10_buf5120}
\end{subfigure}
\caption{
Change in final average accuracy according to the masking value.
All experiments were conducted using 10 trials with random seeds.
The mean and standard deviation of final average accuracy are represented as line and band, respectively.
($\mathcal{B}$: buffer size, {\color{gray} dotted line}: the final average accuracy of the baseline - DER++)
}
\label{fig:analysis:mask_variants}
\end{figure}

\subsection{Ablation Study for Masking Values}
\label{subsec:ablation}

\paragraph{Implementation Details.}
To validate that the impact of general masked softmax is indeed regulated by controlling the masking value, we conducted experiments on split MNIST and CIFAR10 datasets with buffer sizes of 200, 500, and 5120, using various masking value $m$, where $m \in \{\text{-}\infty, \text{-}10^9, \text{-}10^6, \text{-}10^3, \text{-}10, \text{-}5, \text{-}2, \text{-}1, 0\}$. 
All experimental settings, except for the masking value, were identical to those in continual learning benchmarks.

\paragraph{Results and Analyses.}
As illustrated in~\autoref{fig:analysis:seq-cifar10_buf200}, masked softmax with dark knowledge outperforms DER++ in all masking values while MSDER++ shows inferior performance in all masking values, as shown in Figures~\ref{fig:analysis:seq-mnist_buf500} and~\ref{fig:analysis:seq-mnist_buf5120}. 
However, even in these inferior cases, the performance of the model with masked softmax increases as the scale of the masking value decreases. 
This indicates that controlling the masking value is effective in adjusting the \textit{push} effect of softmax.
Furthermore, as illustrated in Figures~\ref{fig:analysis:seq-mnist_buf200},~\ref{fig:analysis:seq-cifar10_buf500}, and~\ref{fig:analysis:seq-cifar10_buf5120}, the model using masked softmax outperforms DER++ in certain masking values. 
This means that by identifying the appropriate masking value for each setting, the model's performance in continual learning can be improved.

%
%
\section{Conclusion and Future Work}
\label{section: conclusion}
In this paper, we shed light on the \textit{pull} and \textit{push} effects of the softmax function when used with cross-entropy loss in continual learning settings. 
The \textit{push} effect of softmax exacerbates catastrophic forgetting by flowing gradients toward the class weight vectors in previous tasks. 
To address this issue, we revisit a well-known softmax masking approach to investigate maintaining confidence against the \textit{push} effect of softmax. 
The use of negative infinity masked softmax has been found to be effective in achieving the desired purpose by producing zero gradients on both old and new classes, thereby increasing model stability. 
However, it has been observed that this approach exhibits inferior performance when transferring previous knowledge to the current task. 
To address the issue, we suggest a \textit{general masked softmax} that regulates the push effect of softmax. 
This is achieved by setting the mask value to negative infinity or real values, while also preventing the gradient flow to old and new classes. 
The effectiveness of our method in adjusting the trade-off between stability and plasticity of the model in continual learning is demonstrated, and it improves prior replay-based methods in continual learning benchmarks, even when the buffer size of episodic memory is set to extremely small values. 
In future work, this method can be extended to provide more fine control in terms of plasticity to reduce memory reliance more effectively.

{
\small

\bibliography{reference}
\bibliographystyle{unsrt}

}

\newpage

\appendix

\section{Additional Experimental Results}
\label{appx:sec:additional_experimental_results}

\begin{table*}[h!]
\tiny
\centering
\caption{Classification results for standard continual learning benchmarks including the performance of commonly used methods in these benchmarks. All experiments were conducted using 10 trials with random seeds except S-Tiny-ImageNet, where the experiment utilized 5 triasls with random seeds. Best in bold for each buffer setting. (Final Average Accuracy $\uparrow$ (\%): \textit{mean}$_{\textit{std}}$)}
\begin{tabular}{clrrrrrrrr}
    \toprule
    \multirow{2}{*}{$\mathcal{B}$} & 
    \multirow{2}{*}{Method} &
    \multicolumn{2}{c}{S-MNIST} &
    \multicolumn{2}{c}{S-CIFAR-10} &
    \multicolumn{2}{c}{S-CIFAR-100} &
    \multicolumn{2}{c}{S-Tiny-ImageNet} \\
    & & 
    \textit{Class-IL} & \textit{Task-IL} &
    \textit{Class-IL} & \textit{Task-IL} &
    \textit{Class-IL} & \textit{Task-IL} &
    \textit{Class-IL} & \textit{Task-IL} \\
    \midrule
    \multirow{2}{*}{-} & JOINT
    & 95.57$_{0.24}$ & 99.51$_{0.07}$ & 92.20$_{0.15}$ & 98.31$_{0.12}$ 
    & \multicolumn{1}{c}{-} & \multicolumn{1}{c}{-} & 59.99$_{0.19}$ & 82.04$_{0.10}$ \\
    & SGD
    & 19.60$_{0.04}$ & 94.94$_{2.18}$ & 19.62$_{0.05}$ & 61.02$_{3.33}$ 
    & \multicolumn{1}{c}{-} & \multicolumn{1}{c}{-} &  7.92$_{0.26}$ & 18.31$_{0.68}$ \\
    \midrule
    \multirow{4}{*}{-} & oEWC{\scriptsize ~\cite{cil_oEWC}}
    & 20.46$_{1.01}$ & 98.39$_{0.48}$ & 19.49$_{0.12}$ & 68.29$_{3.92}$ 
    & \multicolumn{1}{c}{-} & \multicolumn{1}{c}{-} &  7.58$_{0.10}$ & 19.20$_{0.31}$ \\
    & SI{\scriptsize ~\cite{cil_SI}}
    & 19.27$_{0.30}$ & 96.00$_{2.04}$ & 19.48$_{0.17}$ & 68.05$_{5.91}$ 
    & \multicolumn{1}{c}{-} & \multicolumn{1}{c}{-} &  6.58$_{0.31}$ & 36.32$_{0.13}$ \\
    & LwF{\scriptsize ~\cite{cil_LwF}}
    & 19.62$_{0.01}$ & 94.11$_{3.01}$ & 19.61$_{0.05}$ & 63.29$_{2.35}$ 
    & \multicolumn{1}{c}{-} & \multicolumn{1}{c}{-} &  8.46$_{0.22}$ & 15.85$_{0.58}$ \\
    & PNN{\scriptsize ~\cite{cil_PNN}}
    & \multicolumn{1}{c}{-} & 99.23$_{0.20}$ & \multicolumn{1}{c}{-} & 95.13$_{0.72}$ 
    & \multicolumn{1}{c}{-} & \multicolumn{1}{c}{-} & \multicolumn{1}{c}{-} & 67.84$_{0.29}$ \\
    \midrule
    \multirow{16}{*}{200}
    & GEM{\scriptsize ~\cite{cil_replay_gem}}
    & 80.11$_{1.54}$ & 97.78$_{0.25}$ & 25.54$_{0.76}$ & 90.44$_{0.94}$ 
    & \multicolumn{1}{c}{-} & \multicolumn{1}{c}{-} & \multicolumn{1}{c}{-} & \multicolumn{1}{c}{-} \\
    & A-GEM{\scriptsize ~\cite{cil_replay_a-gem}}
    & 45.72$_{4.26}$ & 98.61$_{0.24}$ & 20.04$_{0.34}$ & 83.88$_{1.49}$ 
    & \multicolumn{1}{c}{-} & \multicolumn{1}{c}{-} &  8.07$_{0.08}$ & 22.77$_{0.03}$ \\
    & iCaRL{\scriptsize ~\cite{cil_replay_icarl}}
    & 70.51$_{0.53}$ & 98.28$_{0.09}$ & 49.02$_{3.20}$ & 88.99$_{2.13}$ 
    & \multicolumn{1}{c}{-} & \multicolumn{1}{c}{-} &  7.53$_{0.79}$ & 28.19$_{1.47}$ \\
    & FDR{\scriptsize ~\cite{cil_replay_fdr}}
    & 79.43$_{3.26}$ & 97.66$_{0.18}$ & 30.91$_{2.74}$ & 91.01$_{0.68}$ 
    & \multicolumn{1}{c}{-} & \multicolumn{1}{c}{-} &  8.70$_{0.19}$ & 40.36$_{0.68}$ \\
    & GSS{\scriptsize ~\cite{cil_replay_gss}}
    & 38.90$_{2.49}$ & 95.02$_{1.85}$ & 39.07$_{5.59}$ & 88.80$_{2.89}$ 
    & \multicolumn{1}{c}{-} & \multicolumn{1}{c}{-} & \multicolumn{1}{c}{-} & \multicolumn{1}{c}{-}  \\
    & HAL{\scriptsize ~\cite{cil_replay_hal}}
    & 84.70$_{0.87}$ & 97.96$_{0.21}$ & 32.36$_{2.70}$ & 82.51$_{3.20}$ 
    & \multicolumn{1}{c}{-} & \multicolumn{1}{c}{-} & \multicolumn{1}{c}{-} & \multicolumn{1}{c}{-}  \\
    & ER{\scriptsize ~\cite{cil_replay_er}}
    & 80.43$_{1.89}$ & 97.86$_{0.35}$ & 44.79$_{1.86}$ & 91.19$_{0.94}$ 
    & \multicolumn{1}{c}{-} & \multicolumn{1}{c}{-} &  8.49$_{0.16}$ & 38.17$_{2.00}$ \\
    & DER{\scriptsize~\cite{cil_replay_derpp}}
    & 84.55$_{1.64}$ & 98.80$_{0.15}$ & 61.93$_{1.79}$ & 91.40$_{0.92}$ 
    & \multicolumn{1}{c}{-} & \multicolumn{1}{c}{-} & 11.87$_{0.78}$ & 40.22$_{0.67}$ \\
    & DER++{\scriptsize~\cite{cil_replay_derpp}}
    & 85.61$_{1.40}$ & 98.76$_{0.28}$ & 64.88$_{1.17}$ & 91.92$_{0.60}$ 
    & \multicolumn{1}{c}{-} & \multicolumn{1}{c}{-} & 10.96$_{1.17}$ & 40.87$_{1.16}$ \\
    \cmidrule(lr){2-10}
    & ER$^{\dag}$
    & 78.27$_{1.37}$ & 97.73$_{0.26}$ & 49.38$_{2.15}$ & 91.54$_{0.81}$ 
    & 14.88$_{0.47}$ & 66.75$_{1.06}$ &  8.58$_{0.19}$ & 38.39$_{0.72}$ \\
    & MSER($m=\text{-}\infty$)$^{\dag}$
    & 82.98$_{1.03}$ & 98.13$_{0.16}$ & 61.75$_{6.07}$ & 91.39$_{2.13}$
    & 28.51$_{0.44}$ & 68.51$_{0.87}$ & \textbf{15.47}$_{0.67}$ & 44.11$_{0.50}$ \\
    \cmidrule(lr){3-10}
    & DER++$^{\dag}$
    & 85.64$_{1.02}$ & 98.84$_{0.11}$ & 63.67$_{1.01}$ & 91.61$_{0.73}$ 
    & 24.85$_{1.69}$ & 67.69$_{1.39}$ & 11.59$_{1.07}$ & 41.00$_{0.88}$ \\
    & MSDER++($m=\text{-}\infty$)$^{\dag}$
    & 84.45$_{0.88}$ & 99.03$_{0.09}$ & \textbf{66.35}$_{1.52}$ & \textbf{93.17}$_{0.54}$ 
    & 28.57$_{1.11}$ & 74.02$_{0.76}$ & 13.21$_{0.56}$ & 49.75$_{0.99}$ \\
    & MSDER++($m=\text{-}1$)$^{\dag}$
    & \textbf{88.21}$_{0.49}$ & \textbf{99.07}$_{0.12}$ & 64.38$_{3.42}$ & 92.34$_{1.76}$
    & \textbf{28.70}$_{1.35}$ & \textbf{74.33}$_{0.72}$ & 13.25$_{0.51}$ & \textbf{51.24}$_{0.78}$ \\
    \midrule
    \multirow{16}{*}{500} 
    & GEM{\scriptsize ~\cite{cil_replay_gem}}
    & 85.99$_{1.35}$ & 98.71$_{0.20}$ & 26.20$_{1.26}$ & 92.16$_{0.69}$ 
    & \multicolumn{1}{c}{-} & \multicolumn{1}{c}{-} & \multicolumn{1}{c}{-} & \multicolumn{1}{c}{-} \\
    & A-GEM{\scriptsize ~\cite{cil_replay_a-gem}}
    & 46.66$_{5.85}$ & 98.93$_{0.21}$ & 22.67$_{0.57}$ & 89.48$_{1.45}$ 
    & \multicolumn{1}{c}{-} & \multicolumn{1}{c}{-} &  8.06$_{0.04}$ & 25.33$_{0.49}$ \\
    & iCaRL{\scriptsize ~\cite{cil_replay_icarl}}
    & 70.10$_{1.08}$ & 98.32$_{0.07}$ & 47.55$_{3.95}$ & 88.22$_{2.62}$ 
    & \multicolumn{1}{c}{-} & \multicolumn{1}{c}{-} &  9.38$_{1.53}$ & 31.55$_{3.27}$ \\
    & FDR{\scriptsize ~\cite{cil_replay_fdr}}
    & 85.87$_{4.04}$ & 97.54$_{1.90}$ & 28.71$_{3.23}$ & 93.29$_{0.59}$ 
    & \multicolumn{1}{c}{-} & \multicolumn{1}{c}{-} & 10.54$_{0.21}$ & 49.88$_{0.71}$ \\
    & GSS{\scriptsize ~\cite{cil_replay_gss}}
    & 49.76$_{4.73}$ & 97.71$_{0.53}$ & 49.73$_{4.78}$ & 91.02$_{1.57}$ 
    & \multicolumn{1}{c}{-} & \multicolumn{1}{c}{-} & \multicolumn{1}{c}{-} & \multicolumn{1}{c}{-} \\
    & HAL{\scriptsize ~\cite{cil_replay_hal}}
    & 87.21$_{0.49}$ & 98.03$_{0.22}$ & 41.79$_{4.46}$ & 84.54$_{2.36}$ 
    & \multicolumn{1}{c}{-} & \multicolumn{1}{c}{-} & \multicolumn{1}{c}{-} & \multicolumn{1}{c}{-} \\
    & ER{\scriptsize~\cite{cil_replay_er}}
    & 86.12$_{1.89}$ & 99.04$_{0.18}$ & 57.74$_{0.27}$ & 93.61$_{0.27}$ 
    & \multicolumn{1}{c}{-} & \multicolumn{1}{c}{-} &  9.99$_{0.29}$ & 48.64$_{0.46}$ \\
    & DER{\scriptsize~\cite{cil_replay_derpp}}
    & 90.54$_{1.18}$ & 98.84$_{0.13}$ & 70.51$_{1.67}$ & 93.40$_{0.39}$ 
    & \multicolumn{1}{c}{-} & \multicolumn{1}{c}{-} & 17.75$_{1.14}$ & 51.78$_{0.88}$ \\
    & DER++{\scriptsize~\cite{cil_replay_derpp}}
    & 91.00$_{1.49}$ & 98.94$_{0.27}$ & 72.70$_{1.36}$ & 93.88$_{0.50}$ 
    & \multicolumn{1}{c}{-} & \multicolumn{1}{c}{-} & 19.38$_{1.41}$ & 51.91$_{0.68}$ \\
    \cmidrule(lr){2-10}
    & ER$^{\dag}$
    & 85.99$_{1.52}$ & 99.14$_{0.07}$ & 62.38$_{1.40}$ & 94.12$_{0.31}$ 
    & 21.53$_{0.69}$ & 73.97$_{0.30}$ & 10.12$_{0.22}$ & 48.06$_{0.80}$ \\
    & MSER($m=\text{-}\infty$)$^{\dag}$
    & 89.35$_{0.59}$ & \textbf{99.20}$_{0.16}$ & 70.64$_{1.28}$ & 94.22$_{0.41}$
    & 35.68$_{0.89}$ & 74.77$_{0.71}$ & \textbf{20.43}$_{0.38}$ & 53.21$_{0.84}$ \\
    \cmidrule(lr){3-10}
    & DER++$^{\dag}$
    & \textbf{91.01}$_{0.46}$ & 98.95$_{0.07}$ & 73.15$_{0.80}$ & 94.07$_{0.39}$ 
    & 37.41$_{1.40}$ & 76.07$_{0.60}$ & 19.82$_{0.87}$ & 52.24$_{0.94}$ \\
    & MSDER++($m=\text{-}\infty$)$^{\dag}$
    & 83.10$_{1.22}$ & 99.08$_{0.09}$ & 71.85$_{3.76}$ & 94.28$_{1.49}$ 
    & 37.80$_{0.92}$ & 80.52$_{0.60}$ & 17.71$_{0.58}$ & 59.86$_{1.08}$ \\
    & MSDER++($m=\text{-}1$)$^{\dag}$
    & 88.41$_{1.05}$ & 99.06$_{0.07}$ & \textbf{73.53}$_{0.78}$ & \textbf{94.52}$_{0.47}$
    & \textbf{38.76}$_{1.23}$ & \textbf{80.96}$_{0.41}$ & 17.68$_{0.54}$ & \textbf{60.85}$_{0.91}$ \\
    \midrule
    \multirow{16}{*}{5120} 
    & GEM{\scriptsize ~\cite{cil_replay_gem}}
    & 95.11$_{0.87}$ & 99.44$_{0.12}$ & 25.26$_{3.46}$ & 95.55$_{0.02}$ 
    & \multicolumn{1}{c}{-} & \multicolumn{1}{c}{-} & \multicolumn{1}{c}{-} & \multicolumn{1}{c}{-} \\
    & A-GEM{\scriptsize ~\cite{cil_replay_a-gem}}
    & 54.24$_{6.49}$ & 98.93$_{0.20}$ & 21.99$_{2.29}$ & 90.10$_{2.09}$ 
    & \multicolumn{1}{c}{-} & \multicolumn{1}{c}{-} &  7.96$_{0.13}$ & 26.22$_{0.65}$ \\
    & iCaRL{\scriptsize ~\cite{cil_replay_icarl}}
    & 70.60$_{1.03}$ & 98.32$_{0.11}$ & 55.07$_{1.55}$ & 92.23$_{0.84}$ 
    & \multicolumn{1}{c}{-} & \multicolumn{1}{c}{-} & 14.08$_{1.92}$ & 40.83$_{3.11}$ \\
    & FDR{\scriptsize ~\cite{cil_replay_fdr}}
    & 87.47$_{3.15}$ & 97.79$_{1.33}$ & 19.70$_{0.07}$ & 94.32$_{0.97}$ 
    & \multicolumn{1}{c}{-} & \multicolumn{1}{c}{-} & 28.97$_{0.41}$ & 68.01$_{0.42}$ \\
    & GSS{\scriptsize ~\cite{cil_replay_gss}}
    & 89.39$_{0.75}$ & 98.33$_{0.17}$ & 67.27$_{4.27}$ & 94.19$_{1.15}$ 
    & \multicolumn{1}{c}{-} & \multicolumn{1}{c}{-} & \multicolumn{1}{c}{-} & \multicolumn{1}{c}{-} \\
    & HAL{\scriptsize ~\cite{cil_replay_hal}}
    & 89.52$_{0.96}$ & 98.35$_{0.17}$ & 59.12$_{4.41}$ & 88.51$_{3.32}$ 
    & \multicolumn{1}{c}{-} & \multicolumn{1}{c}{-} & \multicolumn{1}{c}{-} & \multicolumn{1}{c}{-} \\
    & ER{\scriptsize~\cite{cil_replay_er}}
    & 93.40$_{1.29}$ & 99.33$_{0.22}$ & 82.47$_{0.52}$ & 96.98$_{0.17}$ 
    & \multicolumn{1}{c}{-} & \multicolumn{1}{c}{-} & 27.40$_{0.31}$ & 67.29$_{0.23}$ \\
    & DER{\scriptsize~\cite{cil_replay_derpp}}
    & 94.90$_{0.57}$ & 99.29$_{0.11}$ & 83.81$_{0.33}$ & 95.43$_{0.33}$ 
    & \multicolumn{1}{c}{-} & \multicolumn{1}{c}{-} & 36.73$_{0.64}$ & 69.50$_{0.26}$ \\
    & DER++{\scriptsize~\cite{cil_replay_derpp}}
    & 95.30$_{1.20}$ & 99.47$_{0.07}$ & 85.24$_{0.49}$ & 96.12$_{0.21}$
    & \multicolumn{1}{c}{-} & \multicolumn{1}{c}{-} & 39.02$_{0.97}$ & 69.84$_{0.63}$ \\
    \cmidrule(lr){2-10}
    & ER$^{\dag}$
    & 93.42$_{1.08}$ & 99.41$_{0.15}$ & 84.31$_{0.38}$ & \textbf{97.02}$_{0.26}$ 
    & 50.51$_{0.94}$ & 85.53$_{0.73}$ & 27.30$_{0.51}$ & 67.69$_{0.33}$ \\
    & MSER($m=\text{-}\infty$)$^{\dag}$
    & 93.51$_{0.60}$ & 99.38$_{0.12}$ & 82.63$_{1.34}$ & 96.45$_{0.27}$
    & 52.95$_{0.73}$ & 84.20$_{0.58}$ & 35.73$_{0.41}$ & 67.50$_{0.53}$ \\
    \cmidrule(lr){3-10}
    & DER++$^{\dag}$
    & \textbf{95.09}$_{0.56}$ & 99.50$_{0.08}$ & 85.56$_{0.38}$ & 96.30$_{0.22}$ 
    & \textbf{59.62}$_{0.63}$ & 86.61$_{0.32}$ & \textbf{39.66}$_{0.89}$ & 69.95$_{0.32}$ \\
    & MSDER++($m=\text{-}\infty$)$^{\dag}$
    & 93.75$_{0.23}$ & \textbf{99.62}$_{0.05}$ & 84.71$_{0.65}$ & 96.78$_{0.16}$ 
    & 58.18$_{0.43}$ & 87.97$_{0.33}$ & 34.72$_{0.46}$ & 72.40$_{0.25}$ \\
    & MSDER++($m=\text{-}1$)$^{\dag}$
    & 94.36$_{0.22}$ & 99.57$_{0.07}$ & \textbf{85.66}$_{0.54}$ & 96.91$_{0.16}$
    & 59.09$_{0.38}$ & \textbf{88.41}$_{0.22}$ & 35.30$_{0.31}$ & \textbf{73.11}$_{0.12}$ \\
    \bottomrule
\end{tabular}
\label{appx:tab:experiment_contid_learning_avg_acc}
\end{table*}

\begin{table*}[t]
\tiny
\centering
\caption{Classification results for standard continual learning benchmarks including the performance of commonly used methods in these benchmarks. All experiments were conducted using 10 trials with random seeds except S-Tiny-ImageNet, where the experiment utilized 5 trials with random seeds. Best in bold for each buffer setting. (Final Average Forgetting $\downarrow$ (\%): \textit{mean}$_{\textit{std}}$))}
\begin{tabular}{clrrrrrrrr}
    \toprule
    \multirow{2}{*}{$\mathcal{B}$} & 
    \multirow{2}{*}{Method} &
    \multicolumn{2}{c}{S-MNIST} &
    \multicolumn{2}{c}{S-CIFAR-10} &
    \multicolumn{2}{c}{S-CIFAR-100} &
    \multicolumn{2}{c}{S-Tiny-ImageNet} \\
    & & 
    \textit{Class-IL} & \textit{Task-IL} &
    \textit{Class-IL} & \textit{Task-IL} &
    \textit{Class-IL} & \textit{Task-IL} &
    \textit{Class-IL} & \textit{Task-IL} \\
    \midrule
    - & SGD
    & 99.10$_{0.55}$ &  5.15$_{2.74}$ & 96.39$_{0.12}$ & 46.24$_{2.12}$ & \multicolumn{1}{c}{-} & \multicolumn{1}{c}{-} & \multicolumn{1}{c}{-} & \multicolumn{1}{c}{-} \\
    \midrule
    \multirow{4}{*}{-} & oEWC{\scriptsize ~\cite{cil_oEWC}}
    & 97.79$_{1.24}$ &  0.44$_{0.16}$ & 91.64$_{3.07}$ & 29.33$_{3.84}$ & \multicolumn{1}{c}{-} & \multicolumn{1}{c}{-} & \multicolumn{1}{c}{-} & \multicolumn{1}{c}{-} \\
    & SI{\scriptsize ~\cite{cil_SI}}
    & 98.89$_{0.86}$ &  5.15$_{2.74}$ & 95.78$_{0.64}$ & 38.76$_{0.89}$ & \multicolumn{1}{c}{-} & \multicolumn{1}{c}{-} & \multicolumn{1}{c}{-} & \multicolumn{1}{c}{-} \\
    & LwF{\scriptsize ~\cite{cil_LwF}}
    & 99.30$_{0.11}$ &  5.15$_{2.74}$ & 96.69$_{0.25}$ & 32.56$_{0.56}$ & \multicolumn{1}{c}{-} & \multicolumn{1}{c}{-} & \multicolumn{1}{c}{-} & \multicolumn{1}{c}{-} \\
    & PNN{\scriptsize ~\cite{cil_PNN}}
    & -              &  0.00$_{0.00}$ & -              &  0.00$_{0.00}$ & \multicolumn{1}{c}{-} & \multicolumn{1}{c}{-} & \multicolumn{1}{c}{-} & \multicolumn{1}{c}{-} \\
    \midrule
    \multirow{16}{*}{200} 
    & GEM{\scriptsize ~\cite{cil_replay_gem}}
    & 22.32$_{2.04}$ &  1.19$_{0.38}$ & 82.61$_{1.60}$ &  9.27$_{2.07}$ & \multicolumn{1}{c}{-} & \multicolumn{1}{c}{-} & \multicolumn{1}{c}{-} & \multicolumn{1}{c}{-} \\
    & A-GEM{\scriptsize ~\cite{cil_replay_a-gem}}
    & 66.15$_{6.84}$ &  0.96$_{0.28}$ & 95.73$_{0.20}$ & 16.39$_{0.86}$ & \multicolumn{1}{c}{-} & \multicolumn{1}{c}{-} & \multicolumn{1}{c}{-} & \multicolumn{1}{c}{-} \\
    & iCaRL{\scriptsize ~\cite{cil_replay_icarl}}
    & 11.73$_{0.73}$ &  0.28$_{0.08}$ & 28.72$_{0.49}$ &  2.63$_{3.48}$ & \multicolumn{1}{c}{-} & \multicolumn{1}{c}{-} & \multicolumn{1}{c}{-} & \multicolumn{1}{c}{-} \\
    & FDR{\scriptsize ~\cite{cil_replay_fdr}}
    & 21.15$_{4.18}$ &  0.52$_{0.18}$ & 86.40$_{2.67}$ &  7.36$_{0.03}$ & \multicolumn{1}{c}{-} & \multicolumn{1}{c}{-} & \multicolumn{1}{c}{-} & \multicolumn{1}{c}{-} \\
    & GSS{\scriptsize ~\cite{cil_replay_gss}}
    & 74.10$_{3.03}$ &  4.30$_{2.31}$ & 75.25$_{4.07}$ &  8.56$_{1.78}$ & \multicolumn{1}{c}{-} & \multicolumn{1}{c}{-} & \multicolumn{1}{c}{-} & \multicolumn{1}{c}{-} \\
    & HAL{\scriptsize ~\cite{cil_replay_hal}}
    & 14.54$_{1.49}$ &  0.53$_{0.19}$ & 69.11$_{4.21}$ & 12.26$_{0.02}$ & \multicolumn{1}{c}{-} & \multicolumn{1}{c}{-} & \multicolumn{1}{c}{-} & \multicolumn{1}{c}{-} \\
    & ER{\scriptsize~\cite{cil_replay_er}}
    & 21.36$_{2.46}$ &  0.84$_{0.41}$ & 61.24$_{2.62}$ &  7.08$_{0.64}$
    & \multicolumn{1}{c}{-} & \multicolumn{1}{c}{-} 
    & \multicolumn{1}{c}{-} & \multicolumn{1}{c}{-} \\
    & DER{\scriptsize~\cite{cil_replay_derpp}}
    & 17.66$_{2.10}$ &  0.57$_{0.18}$ & 40.76$_{0.42}$ &  6.57$_{0.20}$
    & \multicolumn{1}{c}{-} & \multicolumn{1}{c}{-} 
    & \multicolumn{1}{c}{-} & \multicolumn{1}{c}{-} \\
    & DER++{\scriptsize~\cite{cil_replay_derpp}}
    & 16.27$_{1.73}$ &  0.66$_{0.28}$ & 32.59$_{2.32}$ &  5.16$_{0.21}$
    & \multicolumn{1}{c}{-} & \multicolumn{1}{c}{-} 
    & \multicolumn{1}{c}{-} & \multicolumn{1}{c}{-} \\
    \cmidrule(lr){2-10}
    & ER$^{\dag}$
    & 24.09$_{1.61}$ &  0.89$_{0.24}$ & 58.97$_{2.70}$ &  6.49$_{0.90}$ 
    & 81.61$_{0.61}$ & 24.18$_{1.12}$ & 76.62$_{0.48}$ & 43.52$_{0.85}$ \\
    & MSER($m=\text{-}\infty$)$^{\dag}$
    & 10.22$_{1.20}$ &  0.69$_{0.12}$ & 26.30$_{6.18}$ &  6.18$_{1.56}$
    & \textbf{46.51}$_{1.36}$ & 20.92$_{0.79}$ & \textbf{50.86}$_{0.79}$ & 34.97$_{0.61}$ \\
    \cmidrule(lr){3-10}
    & DER++$^{\dag}$
    & 16.34$_{1.23}$ &  0.53$_{0.12}$ & 34.80$_{1.68}$ &  6.45$_{1.07}$ 
    & 68.80$_{2.76}$ & 24.20$_{1.26}$ & 72.77$_{2.13}$ & 41.11$_{0.97}$ \\
    & MSDER++($m=\text{-}\infty$)$^{\dag}$
    &  \textbf{4.29}$_{0.75}$ &  \textbf{0.32}$_{0.09}$ & 23.10$_{1.46}$ &  \textbf{4.82}$_{0.71}$ 
    & 48.52$_{0.62}$ & 17.34$_{0.77}$ & 52.96$_{0.72}$ & 32.20$_{1.21}$ \\
    & MSDER++($m=\text{-}1$)$^{\dag}$
    &  5.00$_{0.51}$ &  \textbf{0.32}$_{0.10}$ & \textbf{22.84}$_{2.00}$ &  5.50$_{1.47}$
    & 48.48$_{1.59}$ & \textbf{17.09}$_{0.77}$ & 55.22$_{0.56}$ & \textbf{30.32}$_{0.93}$ \\
    \midrule
    \multirow{16}{*}{500} 
    & GEM{\scriptsize ~\cite{cil_replay_gem}}
    & 15.57$_{1.77}$ &  0.54$_{0.15}$ & 74.31$_{4.62}$ &  9.12$_{0.21}$ & \multicolumn{1}{c}{-} & \multicolumn{1}{c}{-} & \multicolumn{1}{c}{-} & \multicolumn{1}{c}{-} \\
    & A-GEM{\scriptsize ~\cite{cil_replay_a-gem}}
    & 65.84$_{7.24}$ &  0.64$_{0.20}$ & 94.01$_{1.16}$ & 14.26$_{4.18}$ & \multicolumn{1}{c}{-} & \multicolumn{1}{c}{-} & \multicolumn{1}{c}{-} & \multicolumn{1}{c}{-} \\
    & iCaRL{\scriptsize ~\cite{cil_replay_icarl}}
    & 11.84$_{0.73}$ &  0.30$_{0.09}$ & 25.71$_{1.10}$ &  2.66$_{2.47}$ & \multicolumn{1}{c}{-} & \multicolumn{1}{c}{-} & \multicolumn{1}{c}{-} & \multicolumn{1}{c}{-} \\
    & FDR{\scriptsize ~\cite{cil_replay_fdr}}
    & 13.90$_{5.19}$ &  1.35$_{2.40}$ & 85.62$_{0.36}$ &  4.80$_{0.00}$ & \multicolumn{1}{c}{-} & \multicolumn{1}{c}{-} & \multicolumn{1}{c}{-} & \multicolumn{1}{c}{-} \\
    & GSS{\scriptsize ~\cite{cil_replay_gss}}
    & 60.35$_{6.03}$ &  0.89$_{0.40}$ & 62.88$_{2.67}$ &  7.73$_{3.99}$ & \multicolumn{1}{c}{-} & \multicolumn{1}{c}{-} & \multicolumn{1}{c}{-} & \multicolumn{1}{c}{-} \\
    & HAL{\scriptsize ~\cite{cil_replay_hal}}
    &  9.97$_{1.62}$ &  0.35$_{0.21}$ & 62.21$_{4.34}$ &  5.41$_{1.10}$ & \multicolumn{1}{c}{-} & \multicolumn{1}{c}{-} & \multicolumn{1}{c}{-} & \multicolumn{1}{c}{-} \\
    & ER{\scriptsize~\cite{cil_replay_er}}
    & 15.97$_{2.46}$ &  0.39$_{0.20}$ & 45.35$_{0.07}$ &  3.54$_{0.35}$
    & \multicolumn{1}{c}{-} & \multicolumn{1}{c}{-} 
    & \multicolumn{1}{c}{-} & \multicolumn{1}{c}{-} \\
    & DER{\scriptsize~\cite{cil_replay_derpp}}
    &  9.58$_{1.52}$ &  0.45$_{0.13}$ & 26.74$_{0.15}$ &  4.56$_{0.45}$
    & \multicolumn{1}{c}{-} & \multicolumn{1}{c}{-} 
    & \multicolumn{1}{c}{-} & \multicolumn{1}{c}{-} \\
    & DER++{\scriptsize~\cite{cil_replay_derpp}}
    &  8.85$_{1.86}$ &  0.35$_{0.15}$ & 22.38$_{4.41}$ &  4.66$_{1.15}$
    & \multicolumn{1}{c}{-} & \multicolumn{1}{c}{-} 
    & \multicolumn{1}{c}{-} & \multicolumn{1}{c}{-} \\
    \cmidrule(lr){2-10}
    & ER$^{\dag}$
    & 16.28$_{1.99}$ &  0.38$_{0.11}$ & 42.83$_{1.89}$ &  3.51$_{0.40}$ 
    & 73.90$_{0.82}$ & 16.17$_{0.40}$ & 74.82$_{0.08}$ & 32.17$_{1.01}$ \\
    & MSER($m=\text{-}\infty$)$^{\dag}$
    &  7.18$_{1.37}$ &  0.39$_{0.17}$ & 17.31$_{1.58}$ &  3.09$_{0.36}$
    & 39.74$_{1.36}$ & 13.94$_{0.64}$ & 51.25$_{0.35}$ & 27.73$_{0.80}$ \\
    \cmidrule(lr){3-10}
    & DER++$^{\dag}$
    &  9.00$_{0.50}$ &  0.41$_{0.05}$ & 23.12$_{1.88}$ &  3.56$_{0.38}$ 
    & 52.74$_{2.65}$ & 15.13$_{0.74}$ & 59.04$_{1.27}$ & 27.96$_{0.72}$ \\
    & MSDER++($m=\text{-}\infty$)$^{\dag}$
    &  \textbf{2.07}$_{0.89}$ &  \textbf{0.23}$_{0.09}$ & 16.95$_{3.77}$ &  \textbf{3.03}$_{0.76}$ 
    & 35.98$_{0.79}$ & 10.39$_{0.68}$ & \textbf{47.00}$_{0.75}$ & 21.53$_{1.00}$ \\
    & MSDER++($m=\text{-}1$)$^{\dag}$
    &  4.00$_{0.78}$ &  0.27$_{0.07}$ & \textbf{14.70}$_{1.31}$ &  3.21$_{0.61}$
    & \textbf{35.82}$_{0.99}$ & \textbf{10.00}$_{0.38}$ & 47.25$_{1.04}$ & \textbf{20.20}$_{1.40}$ \\
    \midrule
    \multirow{16}{*}{5120} 
    & GEM{\scriptsize ~\cite{cil_replay_gem}}
    &  4.30$_{1.16}$ &  0.16$_{0.09}$ & 75.27$_{4.41}$ &  6.91$_{2.33}$ & \multicolumn{1}{c}{-} & \multicolumn{1}{c}{-} & \multicolumn{1}{c}{-} & \multicolumn{1}{c}{-} \\
    & A-GEM{\scriptsize ~\cite{cil_replay_a-gem}}
    & 55.10$_{10.79}$ & 0.63$_{0.21}$ & 84.49$_{3.08}$ & 11.36$_{1.68}$ & \multicolumn{1}{c}{-} & \multicolumn{1}{c}{-} & \multicolumn{1}{c}{-} & \multicolumn{1}{c}{-} \\
    & iCaRL{\scriptsize ~\cite{cil_replay_icarl}}
    & 11.64$_{0.72}$ &  0.26$_{0.06}$ & 24.94$_{0.14}$ &  1.59$_{0.57}$ & \multicolumn{1}{c}{-} & \multicolumn{1}{c}{-} & \multicolumn{1}{c}{-} & \multicolumn{1}{c}{-} \\
    & FDR{\scriptsize ~\cite{cil_replay_fdr}}
    & 11.58$_{3.97}$ &  0.95$_{1.61}$ & 96.64$_{0.19}$ &  1.93$_{0.48}$ & \multicolumn{1}{c}{-} & \multicolumn{1}{c}{-} & \multicolumn{1}{c}{-} & \multicolumn{1}{c}{-} \\
    & GSS{\scriptsize ~\cite{cil_replay_gss}}
    &  7.90$_{1.21}$ &  0.18$_{0.11}$ & 58.11$_{9.12}$ &  7.71$_{2.31}$ & \multicolumn{1}{c}{-} & \multicolumn{1}{c}{-} & \multicolumn{1}{c}{-} & \multicolumn{1}{c}{-} \\
    & HAL{\scriptsize ~\cite{cil_replay_hal}}
    &  6.55$_{1.63}$ &  0.13$_{0.07}$ & 27.19$_{7.53}$ &  5.21$_{0.50}$ & \multicolumn{1}{c}{-} & \multicolumn{1}{c}{-} & \multicolumn{1}{c}{-} & \multicolumn{1}{c}{-} \\
    & ER{\scriptsize~\cite{cil_replay_er}}
    &  6.08$_{1.84}$ &  0.25$_{0.23}$ & 13.99$_{0.12}$ &  0.27$_{0.06}$
    & \multicolumn{1}{c}{-} & \multicolumn{1}{c}{-} 
    & \multicolumn{1}{c}{-} & \multicolumn{1}{c}{-} \\
    & DER{\scriptsize~\cite{cil_replay_derpp}}
    &  4.53$_{0.83}$ &  0.32$_{0.08}$ & 10.12$_{0.80}$ &  2.59$_{0.08}$
    & \multicolumn{1}{c}{-} & \multicolumn{1}{c}{-} 
    & \multicolumn{1}{c}{-} & \multicolumn{1}{c}{-} \\
    & DER++{\scriptsize~\cite{cil_replay_derpp}}
    &  4.19$_{1.63}$ &  0.23$_{0.06}$ &  7.27$_{0.84}$ &  1.18$_{0.19}$
    & \multicolumn{1}{c}{-} & \multicolumn{1}{c}{-} 
    & \multicolumn{1}{c}{-} & \multicolumn{1}{c}{-} \\
    \cmidrule(lr){2-10}
    & ER$^{\dag}$
    &  6.26$_{1.56}$ &  0.21$_{0.10}$ & 14.44$_{0.65}$ &  \textbf{0.46}$_{0.21}$ 
    & 39.25$_{0.62}$ &  4.53$_{0.36}$ & 54.54$_{0.47}$ & 11.60$_{0.25}$ \\
    & MSER($m=\text{-}\infty$)$^{\dag}$
    &  3.12$_{1.10}$ &  0.24$_{0.10}$ &  7.57$_{1.31}$ &  0.81$_{0.28}$
    & 22.01$_{0.40}$ &  4.10$_{0.35}$ & 30.03$_{0.58}$ &  9.50$_{0.52}$ \\
    \cmidrule(lr){3-10}
    & DER++$^{\dag}$
    &  4.59$_{0.79}$ &  0.24$_{0.10}$ &  7.50$_{0.80}$ &  1.05$_{0.34}$ 
    & 25.28$_{0.71}$ &  4.56$_{0.50}$ & 31.60$_{1.79}$ &  9.63$_{0.53}$ \\
    & MSDER++($m=\text{-}\infty$)$^{\dag}$
    &  \textbf{0.90}$_{0.18}$ &  \textbf{0.09}$_{0.06}$ &  \textbf{5.02}$_{0.69}$ &  0.71$_{0.27}$ 
    & \textbf{14.79}$_{0.83}$ &  2.90$_{0.44}$ & \textbf{20.30}$_{0.48}$ &  7.23$_{0.20}$ \\
    & MSDER++($m=\text{-}1$)$^{\dag}$
    &  1.20$_{0.31}$ &  0.13$_{0.05}$ &  5.43$_{0.56}$ &  0.51$_{0.14}$
    & 15.34$_{0.52}$ &  \textbf{2.60}$_{0.27}$ & 21.39$_{0.74}$ &  \textbf{6.86}$_{0.35}$ \\
    \bottomrule
\end{tabular}
\label{appx:tab:experiment_contid_learning_avg_forget}
\end{table*}


\end{document}